\documentclass[11pt,a4paper]{article}
\usepackage[hyperref]{emnlp2020}
\usepackage{times}
\usepackage{latexsym}
\usepackage{tabularx}
\usepackage{booktabs}
\usepackage{graphicx}
\usepackage{xcolor}
\usepackage{colortbl}
\usepackage{soul}
\usepackage{amsfonts}
\usepackage{stmaryrd}
\usepackage[T1]{fontenc}
\usepackage{enumitem}
\usepackage{afterpage}
\usepackage{amsmath}
\usepackage{amssymb}
\usepackage{multirow} 
\usepackage{balance}
\usepackage{subcaption}
\usepackage{stfloats}
\usepackage{pifont} 
\newcommand{\cmark}{\ding{51}}

\usepackage{rotating}
\usepackage{makecell}
\def\rot{\rotatebox}

\aclfinalcopy

\usepackage{microtype}

\newcommand{\update}[1]{\textcolor{black}{#1}}

\newcommand{\task}{KGC}

\newcommand{\fb}{\textsc{FB15K}}
\newcommand{\wn}{\textsc{WN18}}
\newcommand{\benchmark}{\textsc{CoDEx}}
\newcommand{\benchmarkSmall}{{\textsc{CoDEx-S}}}
\newcommand{\benchmarkMed}{\textsc{CoDEx-M}}
\newcommand{\benchmarkLarge}{\textsc{CoDEx-L}}

\newcommand{\newpar}[1]{\vspace{2mm}\noindent\textbf{#1}}
\newcommand{\rel}[1]{\emph{\nolinkurl{#1}}}

\newcommand*\pct{\scalebox{.9}{\%}}

\title{\benchmark: A Comprehensive Knowledge Graph Completion Benchmark}

\author{
  Tara Safavi \\
  University of Michigan \\
  tsafavi@umich.edu \\ 
  \And
  Danai Koutra \\
  University of Michigan \\
  dkoutra@umich.edu \\ 
}

\date{}

\begin{document}
\maketitle
\begin{abstract}
We present \benchmark{}, a set of knowledge graph \textbf{\textsc{Co}}mpletion \textbf{D}atasets \textbf{\textsc{Ex}}tracted from Wikidata and Wikipedia that improve upon existing knowledge graph completion  benchmarks in scope and level of difficulty.
In terms of scope, \benchmark{} comprises three knowledge graphs varying in size and structure, multilingual descriptions of entities and relations, and tens of thousands of \emph{hard negative} triples that are plausible but verified to be false. 
To characterize \benchmark{}, we contribute thorough empirical analyses and benchmarking experiments. 
First, we analyze each \benchmark{} dataset in terms of logical relation patterns. 
Next, we report baseline link prediction and triple classification results  on \benchmark{} for five extensively tuned embedding models. 
Finally, we differentiate \benchmark{} from the popular \fb-237 knowledge graph completion dataset by showing that \benchmark{} covers more diverse and interpretable content, and is a more difficult link prediction benchmark. 
Data, code, and pretrained models are available at \url{https://bit.ly/2EPbrJs}.
\end{abstract}

\section{Introduction}
\label{sec:intro}
Knowledge graphs are multi-relational graphs that express facts about the world by connecting entities (people, places, things, concepts) via different types of relationships. 
The field of automatic knowledge graph completion (\textbf{\task}), which is motivated by the fact that knowledge graphs are usually incomplete, is an active research direction spanning several subfields of artificial intelligence~\cite{nickel2015review,wang2017knowledge,ji2020survey}. 

As progress in artificial intelligence depends heavily on data, a relevant and high-quality benchmark is imperative to evaluating and advancing the state of the art in \task. 
However, the field has largely remained static in this regard over the past decade.
Outdated subsets of Freebase~\cite{bollacker2008freebase} are most commonly used for evaluation in \task{}, even though Freebase had known quality issues~\cite{tanon2016freebase} and was eventually deprecated in favor of the more recent Wikidata knowledge base~\cite{vrandevcic2014wikidata}. 

Indeed, \task{} benchmarks extracted from Freebase like \fb{} and \fb-237~\cite{bordes2013translating,toutanova2015observed} are questionable in quality. 
For example, \fb{} was shown to have train/test leakage~\cite{toutanova2015observed}.
Later in this paper (\S~\ref{sec:comparison-difficulty}), we will show that a relatively large proportion of relations in \fb-237 can be covered by a trivial frequency rule.

To address the need for a solid benchmark in \task{}, we present \textbf{\benchmark}, a set of knowledge graph \textbf{\textsc{Co}}mpletion \textbf{D}atasets \textbf{\textsc{Ex}}tracted from Wikidata and its sister project Wikipedia.
Inasmuch as Wikidata is considered the successor of Freebase, \benchmark{} improves upon existing Freebase-based \task{} benchmarks in terms of scope and level of difficulty (Table~\ref{table:comparison}). 
Our contributions include:

\paragraph{Foundations}
We survey evaluation datasets in encyclopedic knowledge graph completion to motivate a new benchmark (\S~\ref{sec:related} and Appendix~\ref{appx:related}). 

\paragraph{Data}
We introduce \benchmark{}, a benchmark consisting of three knowledge graphs varying in size and structure, entity types, multilingual labels and descriptions, and---unique to \benchmark---manually verified \emph{hard negative} triples (\S~\ref{sec:benchmark}).
To better understand \benchmark{}, we analyze the logical relation patterns in each of its datasets (\S~\ref{sec:pattern-analysis}). 

\paragraph{Benchmarking}
We conduct large-scale model selection and benchmarking experiments, reporting baseline link prediction and triple classification results on \benchmark{} for five widely used embedding models from different architectural classes (\S~\ref{sec:benchmarking}). 

\paragraph{Comparative analysis}
Finally, to demonstrate the unique value of \benchmark{}, we differentiate \benchmark{} from \fb-237 in terms of both content and difficulty (\S~\ref{sec:comparison}).
We show that \benchmark{} covers more diverse and interpretable content, and is a more challenging link prediction benchmark.

\section{Existing datasets}
\label{sec:related}
\begin{table*}[t!]
\centering
\caption{Qualitative comparison of \benchmark{} datasets to existing Freebase-based \task{} datasets (\S~\ref{sec:fb}).
}
\label{table:comparison}
\def\arraystretch{1.1}
\resizebox{\textwidth}{!}{
    \begin{tabular}{ l p{7.8cm} p{7cm} }
    \toprule
    & Freebase variants (\fb, \fb-237) & \cellcolor{gray!15} \benchmark{} datasets \\ 
    \toprule
    Scope (domains) & Multi-domain, with a strong focus on awards, entertainment, and sports (\S~\ref{sec:comparison-content} and Appendix~\ref{appx:lp-comparison}) & \cellcolor{gray!15} Multi-domain, with focuses on writing, entertainment, music, politics, journalism, academics, and science (\S~\ref{sec:comparison-content} and Appendix~\ref{appx:lp-comparison}) \\ 
    \midrule
    Scope (auxiliary data) & Various decentralized versions of \fb{} with, e.g., entity types~\cite{xie2016representation},  sampled negatives~\cite{socher2013reasoning}, and more (Table~\ref{table:lit-review}) & \cellcolor{gray!15} Centralized repository of three datasets with entity types, multilingual text, and manually annotated hard negatives (\S~\ref{sec:benchmark}) \\ 
    \midrule
    Level of difficulty & \fb{} has severe train/test leakage from inverse relations~\cite{toutanova2015observed}; while removal of inverse relations makes \fb-237 harder than \fb{}, \fb-237 still has a high proportion of easy-to-predict relational patterns (\S~\ref{sec:comparison-difficulty}) 
    & \cellcolor{gray!15} Inverse relations removed from all datasets to avoid train/test leakage (\S~\ref{sec:filtering-data-collection}); manually annotated hard negatives for the task of triple classification (\S~\ref{sec:negative-data-collection}); few trivial patterns for the task of link prediction (\S~\ref{sec:comparison-difficulty}) \\ 
    \bottomrule
\end{tabular}
}
\end{table*}

We begin by surveying existing \task{} benchmarks. 
Table~\ref{table:lit-review} in Appendix~\ref{appx:related} provides an overview of evaluation datasets and tasks on a \emph{per-paper} basis across the artificial intelligence, machine learning, and natural language processing communities. 

Note that we focus on \emph{data} rather than \emph{models}, so we only overview relevant evaluation benchmarks here. 
For more on existing \task{} models, both neural and symbolic, we refer the reader to~\cite{meilicke2018fine} and~\cite{ji2020survey}.

\subsection{Freebase extracts}
\label{sec:fb}

These datasets, extracted from the Freebase knowledge graph~\cite{bollacker2008freebase}, are the most popular for \task{} (see Table~\ref{table:lit-review} in Appendix~\ref{appx:related}). 

\newpar{\fb} was introduced by~\citet{bordes2013translating}. 
It contains 14,951 entities, 1,345 relations, and 592,213 triples covering several domains, with a strong focus on awards, entertainment, and sports. 

\newpar{\fb-237} was introduced by~\citet{toutanova2015observed} to remedy data leakage in \fb{}, which contains many test triples that invert triples in the training set. 
\fb-237 contains 14,541 entities, 237 relations, and 310,116 triples.
We compare \fb-237 to \benchmark{} in \S~\ref{sec:comparison} to assess each dataset's content and relative difficulty. 

\subsection{Other encyclopedic datasets}
\label{sec:nell-yago}

\newpar{NELL-995} \cite{xiong2017deeppath} was taken from the Never Ending Language Learner (NELL) system~\cite{mitchell2018never}, which continuously reads the web to obtain and update its knowledge.
NELL-995, a subset of the 995th iteration of NELL, contains 75,492 entities, 200 relations, and 154,213 triples. 
While NELL-995 is general and covers many domains, its mean average precision was less than 50\pct{} around its 1000th iteration~\cite{mitchell2018never}.
A cursory inspection reveals that many of the triples in NELL-995 are nonsensical or overly generic, suggesting that NELL-995 is not a meaningful dataset for \task{} evaluation.\footnote{Some examples:
(\emph{{politician:jobs}}, \emph{{worksfor}}, \emph{{county:god})},
(\emph{person:buddha001}, \emph{parentofperson}, \emph{person:jesus})}

\newpar{YAGO3-10} \cite{dettmers2018convolutional} is a subset of YAGO3~\cite{mahdisoltani2014yago3}, which covers portions of Wikipedia, Wikidata, and WordNet. 
YAGO3-10 has 123,182 entities, 37 relations, and 1,089,040 triples mostly limited to facts about people and locations.
While YAGO3-10 is a high-precision dataset, it was recently shown to be too easy for link prediction because it contains a large proportion of duplicate relations~\cite{akrami2020realistic,pezeshkpour2020revisiting}. 

\subsection{Domain-specific datasets}
\label{sec:domain-specific}

In addition to large encyclopedic knowledge graphs, it is common to evaluate \task{} methods on at least one smaller, domain-specific dataset, typically drawn from the \textbf{WordNet} semantic network~\cite{miller1998wordnet,bordes2013translating}.
Other choices include the Unified Medical Language System (\textbf{UMLS}) database~\cite{mccray2003upper}, the \textbf{Alyawarra kinship} dataset~\cite{kemp2006learning}, the \textbf{Countries} dataset~\cite{bouchard2015approximate}, and variants of a synthetic ``\textbf{family tree}''~\cite{hinton1986learning}. 
As our focus in this paper is encyclopedic knowledge, we do not cover these datasets further. 

\section{Data collection}
\label{sec:benchmark}
\begin{table*}[t!]
\centering
\caption{
\benchmark{} datasets.
(+):~Positive (\emph{true}) triples.
(-):~Verified negative (\emph{false}) triples (\S~\ref{sec:negative-data-collection}).  
We compute multilingual coverage over all labels, descriptions, and entity Wikipedia extracts successfully retrieved for the respective dataset in Arabic (ar), German (de), English (en), Spanish (es), Russian (ru), and Chinese (zh). 
}
\label{table:codex-cores}
\def\arraystretch{1.1}
\resizebox{\textwidth}{!}{
    \begin{tabular}{ l rr r rrrrrr r rrrrrr } 
    \toprule
     & \multirow{2}{*}{$|E|$} & \multirow{2}{*}{$|R|$} & & \multicolumn{5}{c}{Triples $E \times R \times E$} & & \multicolumn{6}{c}{Multilingual coverage} \\
    \cline{5-9}  \cline{11-16}
    &  & & & Train (+) & Valid (+) & Test (+) & Valid (-) & Test (-) & & ar & de & en & es & ru & zh  \\ 
    \toprule
    {\benchmarkSmall{}} & 2,034 & 42 & & 32,888 & 1827 & 1828 & 1827 & 1828 & &  77.38 & 91.87 & 96.38 & 91.55 & 89.17 & 79.36 \\ 
    {\benchmarkMed{}} & 17,050 & 51 & & 185,584 & 10,310 & 10,311 & 10,310 & 10,311 & & 75.80 & 95.20 & 96.95 & 87.91 & 81.88  & 69.63 \\ 
    {\benchmarkLarge{}} & 77,951 & 69 & & 551,193 & 30,622 & 30,622 & - & - & & 67.47 & 90.84 & 92.40 & 81.30 & 71.12 & 61.06 \\ 
    \bottomrule
\end{tabular}
}
\end{table*}

In this section we describe the pipeline used to construct \benchmark{}. 
For reference, we define a knowledge graph $G$ as a multi-relational graph consisting of a set of entities $E$, relations $R$, and factual statements in the form of (\emph{head}, \emph{relation}, \emph{tail}) triples $(h, r, t)  \in E \times R \times E$.

\subsection{Seeding the collection}
\label{sec:main-data-collection}

We collected an initial set of triples using a type of snowball sampling~\cite{goodman1961snowball}.
We first manually defined a broad seed set of entity and relation types common to 13 domains: Business, geography, literature, media and entertainment, medicine, music, news, politics, religion, science, sports, travel, and visual art.
Examples of seed entity types include \emph{airline}, \emph{journalist}, and \emph{religious text}; corresponding seed relation types in each respective domain include \emph{airline alliance}, \emph{notable works}, and \emph{language of work or name}. 
Table~\ref{table:seeds} in Appendix~\ref{appx:seeds} gives all seed entity and relation types.

Using these seeds, we retrieved an initial set of 380,038 entities, 75 relations, and 1,156,222 triples by querying Wikidata for statements of the form (\emph{head entity of seed type}, \emph{seed relation type}, {?}).

\subsection{Filtering the collection}
\label{sec:filtering-data-collection}

To create smaller data snapshots, we filtered the initial 1.15 million triples to $k$-cores, which are maximal subgraphs $G'$ of a given graph $G$ such that every node in $G'$ has a degree of at least $k$~\cite{batagelj2011fast}.\footnote{A similar approach was used to extract the \fb{} dataset from Freebase~\cite{bordes2013translating}.}
We constructed three \benchmark{} datasets  (Table~\ref{table:codex-cores}): 
\begin{itemize}
    \item \textbf{\benchmarkSmall} ($k$ = 15), which has 36k triples.
    Because of its smaller size, we recommend that \benchmarkSmall{} be used for model testing and debugging, as well as evaluation of methods that are less computationally efficient (e.g., symbolic search-based approaches).
    \item \textbf{\benchmarkMed} ($k$ = 10), which has 206k triples. 
    \benchmarkMed{} is all-purpose, being comparable in size to \fb-237 (\S~\ref{sec:fb}), one of the most popular benchmarks for \task{} evaluation. 
    \item \textbf{\benchmarkLarge} ($k$ = 5), which has 612k triples.
    \benchmarkLarge{} is comparable in size to \fb{} (\S~\ref{sec:fb}), and can be used for both general evaluation and ``few-shot'' evaluation. 
\end{itemize}
We also release the raw dump that we collected via snowball sampling, but focus on \benchmark{}-S through L for the remainder of this paper.

To minimize train/test leakage, we removed inverse relations from each dataset~\cite{toutanova2015observed}.
We computed (\emph{head}, \emph{tail}) and (\emph{tail}, \emph{head}) overlap between all pairs of relations, and removed each relation whose entity pair set overlapped with that of another relation more than 50\pct{} of the time. 
Finally, we split each dataset into 90/5/5 train/validation/test triples such that the validation and test sets contained only entities and relations seen in the respective training sets. 

\subsection{Auxiliary information}
\label{sec:aux-data-collection}

An advantage of Wikidata is that it links entities and relations to various sources of rich auxiliary information. 
To enable tasks that involve joint learning over knowledge graph structure and such additional information, we collected: 
\begin{itemize}
    \item \textbf{Entity types} for each entity as given by Wikidata's \emph{instance of} and \emph{subclass of} relations;
    \item \textbf{Wikidata labels and descriptions} for entities, relations, and entity types; and
    \item \textbf{Wikipedia page extracts} (introduction sections) for entities and entity types.
\end{itemize}
For the latter two, we collected text where available in Arabic, German, English, Spanish, Russian, and Chinese. 
We chose these languages because they are all relatively well-represented on Wikidata~\cite{kaffee2017glimpse}.
Table~\ref{table:codex-cores} provides the coverage by language for each \benchmark{} dataset. 

\begin{table*}[t!]
\centering
\caption{Selected examples of hard negatives in \benchmark{} with explanations.
}
\label{table:negative-samples}
\resizebox{\textwidth}{!}{
\begin{tabular}{ ll } 
\toprule
Negative & Explanation \\ 
\toprule
(\emph{Fr\'{e}d\'{e}ric Chopin}, \emph{occupation}, \emph{conductor}) & Chopin was a pianist and a composer, not a conductor. \\
(\emph{Lesotho}, \emph{official language}, \emph{American English}) & English, not American English, is an official language of Lesotho. \\ 
(\emph{Senegal}, \emph{part of}, \emph{Middle East}) & Senegal is part of West Africa. \\ 
\update{(\emph{Simone de Beauvoir}, \emph{field of work}, \emph{astronomy})} & \update{Simone de Beauvoir's field of work was primarily philosophy.} \\ 
(\emph{Vatican City}, \emph{member of}, \emph{UNESCO}) & Vatican City is a UNESCO World Heritage Site but not a member state. \\ 
\bottomrule
\end{tabular}
}
\end{table*}

\subsection{Hard negatives for evaluation}
\label{sec:negative-data-collection}

Knowledge graphs are unique in that they only contain positive statements, meaning that triples \emph{not} observed in a given knowledge graph are not necessarily false, but merely unseen; this is called the Open World Assumption~\cite{galarraga2013amie}. 
However, most machine learning tasks on knowledge graphs require negatives in some capacity.
While different negative sampling strategies exist~\cite{cai2018kbgan}, the most common approach is to randomly perturb observed triples to generate negatives, following~\citet{bordes2013translating}.

While random negative sampling is beneficial and even necessary in the case where a large number of negatives is needed (i.e., training), it is not necessarily useful for evaluation.
For example, in the task of triple classification, the goal is to discriminate between positive (true) and negative (false) triples.
As we show in \S~\ref{sec:tc-results}, triple classification over randomly generated negatives is trivially easy for state-of-the-art models because random negatives are generally not meaningful or plausible. 
Therefore, we generate and manually evaluate \emph{hard negatives} for \task{} evaluation. 

\paragraph{Generation}
To generate hard negatives, we used each pre-trained embedding model from \S~\ref{sec:models} to predict tail entities of triples in \benchmark. 
For each model, we took as candidate negatives the triples $(h, r, \hat{t})$ for which (i)~the type of the predicted tail entity $\hat{t}$ matched the type of the true tail entity $t$; (ii)~$\hat{t}$ was ranked in the top-10 predictions by that model; and (iii)~$(h, r, \hat{t})$ was not observed in $G$.

\paragraph{Annotation}
We manually labeled all candidate negative triples generated for \benchmarkSmall{} and \benchmarkMed{} as \emph{true} or \emph{false} using the guidelines provided in Appendix~\ref{appx:guidelines}.\footnote{We are currently investigating methods for obtaining high-quality crowdsourced annotations of negatives for \benchmarkLarge{}.}
We randomly selected among the triples labeled as \emph{false} to create \textbf{validation and test negatives for \benchmarkSmall{} and \benchmarkMed{}}, examples of which are given in Table~\ref{table:negative-samples}. 
To assess the quality of our annotations, we gathered judgments from two independent native English speakers on a random selection of 100 candidate negatives.
The annotators were provided the instructions from Appendix~\ref{appx:guidelines}. 
On average, our labels agreed with those of the annotators 89.5\pct{} of the time.
Among the disagreements, 81\pct{} of the time we assigned the label \emph{true} whereas the annotator assigned the label \emph{false}, meaning that we were comparatively conservative in labeling negatives.

\section{Analysis of relation patterns}
\label{sec:pattern-analysis}
To give an idea of the types of reasoning necessary for models to perform well on \benchmark, we analyze the presence of learnable binary relation patterns within \benchmark{}.
The three main types of such patterns in knowledge graphs are \textbf{symmetry}, \textbf{inversion}, and \textbf{compositionality}~\cite{trouillon2019inductive,sun2019rotate}.
We address symmetry and compositionality here, and omit inversion because we specifically removed inverse relations to avoid train/test leakage (\S~\ref{sec:filtering-data-collection}). 

\subsection{Symmetry}
\label{sec:symmetry}

Symmetric relations are relations $r$ for which $(h, r, t) \in G$ implies $(t, r, h) \in G$.
For each relation, we compute the number of its (\emph{head}, \emph{tail}) pairs that overlap with its (\emph{tail}, \emph{head}) pairs, divided by the total number of pairs, and take those with 50\pct{} overlap or higher as symmetric. 
\benchmark{} datasets have five such relations: \emph{diplomatic relation}, \emph{shares border with}, \emph{sibling}, \emph{spouse}, and \emph{unmarried partner}. 
Table~\ref{table:patterns} gives the proportion of triples containing symmetric relations per dataset.
Symmetric patterns are more prevalent in \benchmarkSmall{}, whereas the larger datasets are mostly \textbf{antisymmetric}, i.e., $(h, r, t) \in G$ implies $(t, r, h) \not\in G$. 

\subsection{Composition}
\label{sec:composition}

Compositionality captures \textbf{path rules} of the form $(h, r_1, x_1), \hdots, (x_n, r_n, t) \rightarrow (h, r, t)$.
To learn these rules, models must be capable of ``multi-hop'' reasoning on knowledge graphs~\cite{guu2015traversing}. 

To identify compositional paths, we use the AMIE3 system~\cite{lajus2020fast}, which outputs rules with confidence scores that capture how many times those rules are seen versus violated, to identify paths of lengths two and three; we omit longer paths as they are relatively costly to compute. 
We identify 26, 44, and 93 rules in \benchmarkSmall{}, \benchmarkMed{}, and \benchmarkLarge{}, respectively, with average confidence (out of 1) of 0.630, 0.556, and 0.459.
Table~\ref{table:patterns} gives the percentage of triples per dataset participating in a discovered rule.

Evidently, composition is especially prevalent in \benchmarkLarge{}.
An example rule in \benchmarkLarge{} is ``if \emph{X} was founded by \emph{Y}, and \emph{Y}'s country of citizenship is \emph{Z}, then the country [i.e., of origin] of \emph{X} is \emph{Z}'' (confidence 0.709). 
We release these rules as part of \benchmark{} for further development of \task{} methodologies that incorporate or learn rules. 

\begin{table}[t!]
\centering
\caption{Relation patterns in \benchmark{}. 
For symmetry, we give the proportion of triples containing a symmetric relation. 
For composition, we give the proportion of triples participating in a rule of length two or three. 
}
\label{table:patterns}
\resizebox{\columnwidth}{!}{
    \begin{tabular}{l r r r}
    \toprule 
    & \benchmarkSmall{} & \benchmarkMed{} & \benchmarkLarge{} \\ 
    \toprule 
    Symmetry & 17.46\pct{} & 4.01\pct{} & 3.29\pct{} \\ 
    Composition & 10.09\pct{} & 16.55\pct{} & 31.84\pct{} \\ 
    \bottomrule
\end{tabular}
}
\end{table}

\section{Benchmarking}
\label{sec:benchmarking}
Next, we benchmark performance on \benchmark{} for the tasks of link prediction and triple classification. 
To ensure that models are fairly and accurately compared, we follow~\citet{ruffinelli2020you}, who conducted what is (to the best of our knowledge) the largest-scale hyperparameter tuning study of knowledge graph embeddings to date. 

Note that \benchmark{} can be used to evaluate any type of \task{} method.
However, we focus on embeddings in this section due to their widespread usage in modern NLP~\cite{ji2020survey}. 

\subsection{Tasks}
\label{sec:eval-tasks}

\paragraph{Link prediction}
The link prediction task is conducted as follows: 
Given a test triple $(h, r, t)$, we construct queries $(?, r, t)$ and $(h, r, ?)$.
For each query, a model scores candidate head (tail) entities $\hat{h}$ ($\hat{t}$) according to its belief that $\hat{h}$ ($\hat{t}$) completes the triple (i.e., answers the query). 
The goal is of link prediction is to rank true triples $(\hat{h}, r, t)$ or $(h, r, \hat{t})$ higher than false and unseen triples. 

Link prediction performance is evaluated with mean reciprocal rank (\textbf{MRR}) and \textbf{hits@$k$}.
MRR is the average reciprocal of each ground-truth entity's rank over all $(?, r, t)$ and $(h, r, ?)$ test triples.
Hits@$k$ measures the proportion of test triples for which the ground-truth entity is ranked in the top-$k$ predicted entities. 
In computing these metrics, we exclude the predicted entities for which $(\hat{h}, r, t) \in G$ or $(h, r, \hat{t}) \in G$ so that known positive triples do not artificially lower ranking scores.
This is called ``filtering''~\cite{bordes2013translating}.

\paragraph{Triple classification}
Given a triple $(h, r, t)$, the goal of triple classification is to predict a corresponding label $y \in \{-1, 1\}$.
Since knowledge graph embedding models output real-valued scores for triples, we convert these scores into labels by selecting a decision threshold per relation on the validation set such that validation accuracy is maximized for the model in question.
A similar approach was used by~\citet{socher2013reasoning}.

We compare results on three sets of evaluation negatives: 
(1)~We generate one negative per positive by replacing the positive triple's tail entity by a tail entity $t'$ sampled \textbf{uniformly at random};
(2)~We generate negatives by sampling tail entities according to their \textbf{relative frequency in the tail slot} of all triples; and
(3)~We use the \benchmark{} \textbf{hard negatives}. 
We measure \textbf{accuracy} and \textbf{F1 score}. 

\subsection{Models}
\label{sec:models}

We compare the following embedding methods: 
\textbf{RESCAL} \cite{nickel2011three},
\textbf{TransE} \cite{bordes2013translating}, 
\textbf{ComplEx} \cite{trouillon2016complex},
\textbf{ConvE} \cite{dettmers2018convolutional}, and \textbf{TuckER}~\cite{balazevic2019tucker}. 
These models represent several classes of architecture, from linear (RESCAL, TuckER, ComplEx) to translational (TransE) to nonlinear/learned (ConvE). 
Appendix~\ref{appx:models} provides more specifics on each model. 

\begin{table*}[t!]
\centering
\caption{Comparison of link prediction performance on \benchmark.
}
\label{table:lp-results}
\resizebox{0.85\textwidth}{!}{
    \begin{tabular}{ l c c c c c c c c c c c  } 
    \toprule
     & \multicolumn{3}{c}{\benchmarkSmall{}} & & \multicolumn{3}{c}{\benchmarkMed{}} & & \multicolumn{3}{c}{\benchmarkLarge{}} \\ 
     \cline{2-4} \cline{6-8} \cline{10-12}
     & MRR & Hits@1 & Hits@10 &  & MRR & Hits@1 & Hits@10 &  & MRR & Hits@1 & Hits@10 \\ 
    \toprule
    RESCAL & 0.404 & 0.293 & 0.623 && 0.317 & 0.244 & 0.456 && 0.304 & 0.242 & 0.419 \\ 
    TransE & 0.354 & 0.219 & 0.634 && 0.303 & 0.223 & 0.454 & & 0.187 & 0.116 & 0.317 \\ 
    ComplEx & \textbf{0.465} & \textbf{0.372} & \textbf{0.646} && \textbf{0.337} & \textbf{0.262} & \textbf{0.476} & & 0.294 & 0.237 & 0.400 \\
    ConvE & 0.444 & 0.343 & 0.635 && 0.318 & 0.239 & 0.464 & & 0.303 & 0.240 & 0.420  \\
    TuckER & 0.444 & 0.339 & 0.638 && 0.328 & 0.259 & 0.458 & &  \textbf{0.309} & \textbf{0.244} & \textbf{0.430} \\ 
    \bottomrule
\end{tabular}
}
\end{table*}

\subsection{Model selection}
\label{sec:model-selection}

As recent studies have observed that training strategies are equally, if not more, important than architecture for link prediction~\cite{kadlec2017knowledge,lacroix2018canonical,ruffinelli2020you}, we search across a large range of hyperparameters to ensure a truly fair comparison. 
To this end we use the PyTorch-based LibKGE framework for training and selecting knowledge graph embeddings.\footnote{\url{https://github.com/uma-pi1/kge}}
In the remainder of this section we outline the most important parameters of our model selection process. 
Table~\ref{table:hyperparameter-search-space} in Appendix~\ref{appx:hyperparameters} gives further details and all hyperparameter ranges and values. 
All experiments were run on a single NVIDIA Tesla V100 GPU with 16 GB of RAM. 

\paragraph{Training negatives}
Given a set of positive training triples $\{(h, r, t)\}$, 
we compare three types of negative sampling strategy implemented by LibKGE: 
(a)~\textbf{NegSamp}, or randomly corrupting head entities $h$ or tail entities $t$ to create negatives;
(b)~\textbf{1vsAll}, or treating \emph{all} possible head/tail corruptions of $(h, r, t)$ as negatives, including the corruptions that are actually positives; and
(c)~\textbf{KvsAll}, or treating batches of head/tail corruptions \emph{not} seen in the knowledge graph as negatives. 

\paragraph{Loss functions}
We consider the following loss functions:
(i)~\textbf{MR} or margin ranking, which aims to maximize a margin between positive and negative triples;
(ii)~\textbf{BCE} or binary cross-entropy, which is computed by applying the logistic sigmoid to triple scores; and
(iii)~\textbf{CE} or cross-entropy between the softmax over the entire distribution of triple scores and the label distribution over all triples, normalized to sum to one. 

\paragraph{Search strategies}
We select models using the Ax platform, which supports hyperparameter search using both quasi-random sequences of generated configurations and Bayesian optimization (BO) with Gaussian processes.\footnote{\url{https://ax.dev/}} 
At a high level, for each dataset and model, we generate both quasi-random and BO trials \emph{per negative sampling and loss function combination}, ensuring that we search over a wide range of hyperparameters for different types of training strategy. 
Appendix~\ref{appx:hyperparameters} provides specific details on the search strategy for each dataset, which was determined according to resource constraints and observed performance patterns. 

\begin{table*}[h!]
\centering
    \caption{Comparison of triple classification performance on \benchmark{} by negative generation strategy. 
    }
    \label{table:tc-results}
    \resizebox{\textwidth}{!}{
    \begin{tabular}{ l cccccccc c cccccccc } 
    \toprule
    & \multicolumn{8}{c}{\benchmarkSmall{}} && \multicolumn{8}{c}{\benchmarkMed{}}  \\
    \cline{2-9} \cline{11-18} 
    & \multicolumn{2}{c}{Uniform} && \multicolumn{2}{c}{Relative freq.} && \multicolumn{2}{c}{Hard neg.} && \multicolumn{2}{c}{Uniform} && \multicolumn{2}{c}{Relative freq.} && \multicolumn{2}{c}{Hard neg.} \\ 
    \cline{2-3} \cline{5-6} \cline{8-9} \cline{11-12} \cline{14-15} \cline{17-18}
    & Acc. & F1 && Acc. & F1 && Acc. & F1 && Acc. & F1 && Acc. & F1 && Acc. & F1 \\ 
    \toprule
    RESCAL & 0.972 & 0.972 && 0.916 & 0.920 && \update{\textbf{0.843}} & \update{\textbf{0.852}} && 0.977 & 0.976 && 0.921 & 0.922 && \update{0.818} & \update{0.815} \\
    TransE & 0.974 & 0.974 && 0.919 & 0.923 && \update{0.829} & \update{0.837} && \textbf{0.986} & \textbf{0.986} && 0.932 & 0.933 && \update{0.797} & \update{0.803} \\
    ComplEx & \textbf{0.975} & \textbf{0.975} && \textbf{0.927} & \textbf{0.930} && \update{0.836} & \update{0.846} && 0.984 & 0.984 &&  0.930 & 0.933 && \update{0.824} & \update{0.818}  \\
    ConvE & 0.972 & 0.972 && 0.921 & 0.924 && \update{0.841} & \update{0.846} && 0.979 & 0.979 && \textbf{0.934} & \textbf{0.935}  && \update{\textbf{0.826}} & \update{\textbf{0.829}} \\
    \update{TuckER} & \update{0.973} & \update{0.973} && \update{0.917} & \update{0.920} && \update{0.840} & \update{0.846} && \update{0.977} & \update{0.977} &&  \update{0.920} & \update{0.922} && \update{0.823} & \update{0.816} \\
    \bottomrule
    \end{tabular}
}
\end{table*}

\subsection{Link prediction results}
\label{sec:lp-results}

Table~\ref{table:lp-results} gives link prediction results. 
We find that ComplEx is the \textbf{best at modeling symmetry and antisymmetry}, \update{and indeed it was designed specifically to improve upon bilinear models that do not capture symmetry, like DistMult~\cite{trouillon2016complex}}. 
As such, it performs the best on \benchmarkSmall{}, which has the highest proportion of symmetric relations.
For example, on the most frequent symmetric relation (\emph{diplomatic relation}),  
ComplEx achieves 0.859 MRR, compared to 0.793 for ConvE, 0.490 for RESCAL, and 0.281 for TransE. 

\begin{figure}[t!]
    \centering
    \includegraphics[width=0.85\columnwidth]{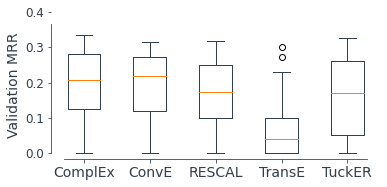}
    \captionof{figure}{Distribution of validation MRR, \benchmarkMed{}.
    }
    \label{fig:valid-boxplot}
\end{figure}

By contrast, TuckER is \textbf{strongest at modeling compositional relations}, so it performs best on \benchmarkLarge{}, which has a high degree of compositionality.
For example, on the most frequent compositional relation in \benchmarkLarge{} (\emph{languages spoken, written, or signed}),
\update{TuckER achieves 0.465 MRR}, compared to 0.464 for RESCAL, 0.463 for ConvE, 0.456 for ComplEx, and  0.385 for TransE.
By contrast, since \benchmarkMed{} is mostly asymmetric and non-compositional, ComplEx performs best because of its ability to model asymmetry. 

\paragraph{Effect of hyperparameters}
As shown by Figure~\ref{fig:valid-boxplot}, hyperparameters have a strong impact on link prediction performance: Validation MRR for all models varies by \emph{over 30 percentage points} depending on the training strategy and input configuration.
This finding is consistent with previous observations in the literature~\cite{kadlec2017knowledge,ruffinelli2020you}. Appendix~\ref{appx:hyperparameters} provides the best configurations for each model.

Overall, we find that the choice of loss function in particular significantly impacts model performance.
Each model consistently \textbf{achieved its respective peak performance with cross-entropy (CE) loss}, a finding which is corroborated by several other \task{} comparison papers~\cite{kadlec2017knowledge,ruffinelli2020you,jain2020knowledge}.  
As far as negative sampling techniques, we do not find that a single strategy is dominant, suggesting that the choice of loss function is more important.

\subsection{Triple classification results}
\label{sec:tc-results}

Table~\ref{table:tc-results} gives triple classification results. 
Evidently, triple classification on \textbf{\emph{randomly} generated negatives is a nearly-solved task}.
On negatives generated uniformly at random, performance scores are nearly identical at almost 100\pct{} accuracy. 
Even with a negative sampling strategy ``smarter'' than uniform random, all models perform well. 

\paragraph{Hard negatives}
Classification performance degenerates considerably on our hard negatives, around 8 to 11 percentage points from relative frequency-based sampling and 13 to 19 percentage points from uniformly random sampling.
Relative performance also varies: 
In contrast to our link prediction task in which ComplEx and TuckER were by far the strongest models, RESCAL is slightly stronger on the \benchmarkSmall{} hard negatives, whereas ConvE performs best on the \benchmarkMed{} hard negatives.
These results indicate that \textbf{triple classification is indeed a distinct task} that requires different architectures and, in many cases, different training strategies (Appendix~\ref{appx:hyperparameters}). 

We believe that few recent works use triple classification as an evaluation task because of the lack of true hard negatives in existing benchmarks.
Early works reported high triple classification accuracy on sampled negatives~\cite{socher2013reasoning,wang2014knowledge}, perhaps leading the community to believe that the task was nearly solved.
However, our results demonstrate that the task is far from solved when the negatives are plausible but truly false.

\section{Comparative case study}
\label{sec:comparison}
Finally, we conduct a comparative analysis between \benchmarkMed{} and \fb-237 (\S~\ref{sec:fb}) to demonstrate the unique value of \benchmark{}. 
We choose \fb-237 because it is the most popular encyclopedic \task{} benchmark after \fb{}, which was already shown to be an easy dataset by~\citet{toutanova2015observed}. 
We choose \benchmarkMed{} because it is the closest in size to \fb-237. 

\begin{figure*}[t!]
    \centering
    \includegraphics[width=\textwidth]{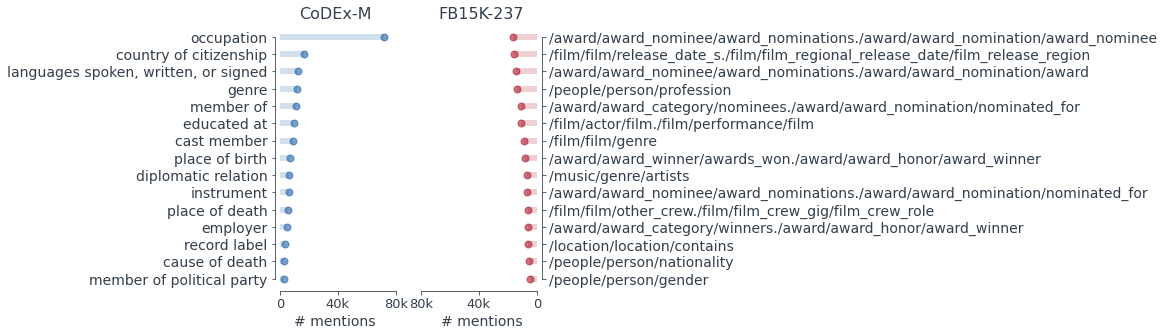}
    \caption{Top-15 most frequent relations in \benchmarkMed{} and \fb-237.
    }
    \label{fig:top-k-relations}
\end{figure*}

\subsection{Content}
\label{sec:comparison-content}

We first compare the content in \benchmarkMed{}, which is extracted from Wikidata, with that of \fb-237, which is extracted from Freebase.
For brevity, Figure~\ref{fig:top-k-relations} compares the top-15 relations by mention count in the two datasets. 
Appendix~\ref{appx:lp-comparison} provides more content comparisons. 

\paragraph{Diversity}
The most common relation in \benchmarkMed{} is \emph{occupation}, which is because most people on Wikidata have multiple occupations listed. 
By contrast, the frequent relations in \fb-237 are mostly related to awards and film.
In fact, over 25\pct{} of all triples in \fb-237 belong to the \emph{/award} relation domain, suggesting that \benchmark{} covers a more diverse selection of content. 

\paragraph{Interpretability}
The Freebase-style relations are also arguably less interpretable than those in Wikidata.
Whereas Wikidata relations have concise natural language labels, the Freebase relation labels are hierarchical, often at five or six levels of hierarchy (Figure~\ref{fig:top-k-relations}).
Moreover, all relations in Wikidata are binary, whereas some Freebase relations are $n$-nary~\cite{tanon2016freebase}, meaning that they connect more than two entities.
The relations containing a dot (``.'') are such $n$-nary relations, and are difficult to reason about without understanding the structure of Freebase, which has been deprecated. 
We further discuss the impact of such $n$-nary relations for link prediction in the following section. 

\subsection{Difficulty}
\label{sec:comparison-difficulty}

Next, we compare the datasets in a link prediction task to show that \benchmarkMed{} is more difficult.

\paragraph{Baseline}
We devise a ``non-learning'' link prediction baseline.
Let $(h, r, ?)$ be a test query.
Our baseline scores candidate tail entities by their relative frequency in the tail slot of all training triples mentioning $r$, filtering out tail entities $t$ for which $(h, r, t)$ is already observed in the training set.
If all tail entities $t$ are filtered out, we score entities by frequency before filtering. 
The logic of our approach works in reverse for $(?, r, t)$ queries.
In evaluating our baseline, we follow LibKGE's protocol for breaking ties in ranking (i.e., for entities that appear with equal frequency) by taking the mean rank of all entities with the same score. 

\paragraph{Setup}
We compare our baseline to the best pre-trained embedding model per dataset: RESCAL for \fb-237, which was released by~\citet{ruffinelli2020you}, and ComplEx for \benchmarkMed{}.
We evaluate performance with MRR and Hits@10. 
Beyond overall performance, we also compute \emph{per-relation improvement} of the respective embedding over our baseline in terms of percentage points MRR. 
This measures the amount of learning beyond frequency statistics necessary for each relation. 

\begin{figure}[t!]
    \centering
    \includegraphics[width=0.9\columnwidth]{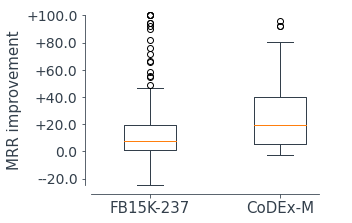}
    \caption{Improvement in MRR of the embedding over our frequency baseline per relation type.
    Negative means that our baseline outperforms the embedding. 
    The medians are 8.27 and 20.04 percentage points on \fb-237 and \benchmarkMed{}, respectively. 
    }
    \label{fig:mrr-diff}
\end{figure}

\paragraph{Results and discussion}
Table~\ref{table:baseline} compares the overall performance of our baseline versus the best embedding per dataset, and 
Figure~\ref{fig:mrr-diff} shows the improvement of the respective embedding over our baseline {per relation type} on each dataset. 
The improvement of the embedding is much smaller on \fb-237 than \benchmarkMed{}, and in fact our baseline performs on par with or even \emph{outperforms} the embedding on \fb-237 for some relation types. 

\begin{table}[t!]
\centering
\caption{Overall performance (MRR) of our frequency baseline versus the best embedding nodel per benchmark. 
``Improvement'' refers to the improvement of the embedding over the baseline. 
}
\label{table:baseline}
\resizebox{0.9\columnwidth}{!}{
    \begin{tabular}{ l r r r  } 
    \toprule
     & Baseline & Embedding & Improvement \\ 
     \toprule
     \fb-237 & 0.236 & 0.356 & +0.120 \\
    \benchmarkMed{} & 0.135 & 0.337 & +0.202 \\
    \bottomrule
\end{tabular}
}
\end{table}

\begin{figure}[t!]
    \centering
    \includegraphics[width=0.99\columnwidth]{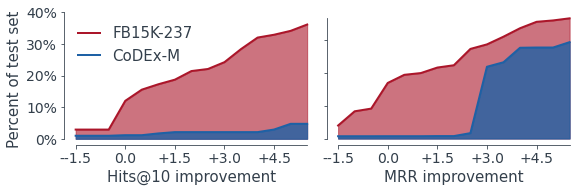}
    \caption{Empirical CDF of improvement of the best embedding over our frequency baseline. 
    }
    \label{fig:ecdf}
\end{figure}

To further explore these cases, Figure~\ref{fig:ecdf} gives the empirical cumulative distribution function  of improvement, which shows the percentage of test triples for which the level of improvement is {less than or equal to a given value} on each dataset.  
Surprisingly, the improvement for both MRR and Hits@10 is less than five percentage points for nearly 40\pct{} of \fb-237's test set, and is \emph{zero or negative} 15\pct{} of the time.
By contrast, our baseline is significantly weaker than the strongest embedding method on \benchmarkMed{}. 

The disparity in improvement is due to two relation patterns prevalent in \fb-237:
\begin{itemize}
    \item \textbf{Skewed relations} \fb-237 contains many relations that are skewed toward a single head or tail entity. 
    For example, our baseline achieves perfect performance over all $(h, r, ?)$ queries for the \rel{/common/topic/webpage./common/webpage/category} relation because this relation has only \emph{one} unique tail entity. 
    Another example of a highly skewed relation in \fb-237 is \rel{/people/person/gender}, for which 78.41\pct{} of tails are the entity \emph{male}. 
    In fact, 11 relations in \fb-237 have only {one} unique tail entity, accounting for 3.22\pct{} of all tail queries in \fb-237. 
    Overall, 15.98\pct{} of test triples in \fb-237 contain relations that are skewed 50\pct{} or more toward a single head or tail entity, whereas only 1.26\pct{} of test triples  in \benchmarkMed{} contain such skewed relations. 
    \item \textbf{Fixed-set relations} Around 12.7\pct{} of test queries in \fb-237 contain relation types that connect entities to \emph{fixed sets of values}.
    As an example, each head entity that participates in the \fb-237 relation \rel{{/travel/travel\_destination/climate./travel/travel\_destination\_monthly_climate/month}} is connected to the \emph{same} 12 tails (months of the year) throughout train, validation, and test.
    This makes prediction trivial with our baseline: By filtering out the tail entities already seen in train, only a few (or even one) candidate tail(s) are left in test, and the answer is guaranteed to be within these candidates.
    These relations only occur in \fb{}-237 because of the way the dataset was constructed from Freebase.
    Specifically, Freebase used a special type of entity called {Compound Value Type} (CVT) as an intermediary node connecting $n$\emph{-ary} relations. 
    Binary relations were created by traversing through CVTs, yielding some relations that connect entities to fixed sets of values. 
\end{itemize}
We conclude that while \fb-237 is a valuable dataset,
\benchmark{} is more appropriately difficult for link prediction.
Additionally, we note that in \fb-237, all validation and test triples containing entity pairs directly linked in the training set were deleted~\cite{toutanova2015observed}, meaning that symmetry cannot be tested for in \fb-237.
Given that \benchmark{} datasets contain both symmetry and compositionality, \benchmark{} is more suitable for assessing how well models can learn relation patterns that go beyond frequency.

\section{Conclusion and outlook}
\label{sec:conclusion}
We present \textbf{\benchmark}, a set of knowledge graph \textbf{\textsc{Co}}mpletion \textbf{D}atasets \textbf{\textsc{Ex}}tracted from Wikidata and Wikipedia, and show that \benchmark{} is suitable for multiple \task{} tasks. 
We release data, code, and pretrained models for use by the community at \url{https://bit.ly/2EPbrJs}. 
Some promising future directions on \benchmark{} include: 
\begin{itemize}
    \item \textbf{Better model understanding}
    \benchmark{} can be used to analyze the impact of hyperparameters, training strategies, and model architectures in \task{} tasks. 
    \item \textbf{Revival of triple classification}
    We encourage the use of triple classification on \benchmark{} in addition to link prediction because it directly tests discriminative power. 
    \item \textbf{Fusing text and structure}
    Including text in both the link prediction and triple classification tasks should substantially improve performance~\cite{toutanova2015representing}. 
    Furthermore, text can be used for few-shot link prediction, an emerging research direction~\cite{xiong2017deeppath,shi2017proje}. 
\end{itemize}
Overall, we hope that \benchmark{} will provide a boost to research in \task, which will in turn impact many other fields of artificial intelligence. 

\section*{Acknowledgments}
The authors thank Michał Rybak and Xinyi (Carol) Zheng for their contributions. 
This material is supported by the National Science Foundation under Grant No. IIS 1845491,  Army Young Investigator Award No. W911NF1810397, and an NSF Graduate Research Fellowship.

\balance
\bibliography{references}

\begin{thebibliography}{74}
\expandafter\ifx\csname natexlab\endcsname\relax\def\natexlab#1{#1}\fi

\bibitem[{Akrami et~al.(2020)Akrami, Saeef, Zhang, Hu, and
  Li}]{akrami2020realistic}
Farahnaz Akrami, Mohammed~Samiul Saeef, Qingheng Zhang, Wei Hu, and Chengkai
  Li. 2020.
\newblock Realistic re-evaluation of knowledge graph completion methods: An
  experimental study.
\newblock In \emph{SIGMOD}.

\bibitem[{Balazevic et~al.(2019{\natexlab{a}})Balazevic, Allen, and
  Hospedales}]{balazevic2019multi}
Ivana Balazevic, Carl Allen, and Timothy Hospedales. 2019{\natexlab{a}}.
\newblock Multi-relational poincar{\'e} graph embeddings.
\newblock In \emph{NeurIPS}.

\bibitem[{Balazevic et~al.(2019{\natexlab{b}})Balazevic, Allen, and
  Hospedales}]{balazevic2019tucker}
Ivana Balazevic, Carl Allen, and Timothy Hospedales. 2019{\natexlab{b}}.
\newblock Tucker: Tensor factorization for knowledge graph completion.
\newblock In \emph{EMNLP-IJCNLP}.

\bibitem[{Bansal et~al.(2019)Bansal, Juan, Ravi, and McCallum}]{bansal2019a2n}
Trapit Bansal, Da-Cheng Juan, Sujith Ravi, and Andrew McCallum. 2019.
\newblock A2n: Attending to neighbors for knowledge graph inference.
\newblock In \emph{ACL}.

\bibitem[{Batagelj and Zaver{\v{s}}nik(2011)}]{batagelj2011fast}
Vladimir Batagelj and Matja{\v{z}} Zaver{\v{s}}nik. 2011.
\newblock Fast algorithms for determining (generalized) core groups in social
  networks.
\newblock \emph{ADAC}, 5(2).

\bibitem[{Bollacker et~al.(2008)Bollacker, Evans, Paritosh, Sturge, and
  Taylor}]{bollacker2008freebase}
Kurt Bollacker, Colin Evans, Praveen Paritosh, Tim Sturge, and Jamie Taylor.
  2008.
\newblock Freebase: a collaboratively created graph database for structuring
  human knowledge.
\newblock In \emph{SIGMOD}.

\bibitem[{Bordes et~al.(2013)Bordes, Usunier, Garcia-Duran, Weston, and
  Yakhnenko}]{bordes2013translating}
Antoine Bordes, Nicolas Usunier, Alberto Garcia-Duran, Jason Weston, and Oksana
  Yakhnenko. 2013.
\newblock Translating embeddings for modeling multi-relational data.
\newblock In \emph{NeurIPS}.

\bibitem[{Bouchard et~al.(2015)Bouchard, Singh, and
  Trouillon}]{bouchard2015approximate}
Guillaume Bouchard, Sameer Singh, and Theo Trouillon. 2015.
\newblock On approximate reasoning capabilities of low-rank vector spaces.
\newblock In \emph{AAAI Spring Symposium Series}.

\bibitem[{Cai and Wang(2018)}]{cai2018kbgan}
Liwei Cai and William~Yang Wang. 2018.
\newblock Kbgan: Adversarial learning for knowledge graph embeddings.
\newblock In \emph{NAACL-HLT}.

\bibitem[{Das et~al.(2018)Das, Dhuliawala, Zaheer, Vilnis, Durugkar,
  Krishnamurthy, Smola, and McCallum}]{das2017go}
Rajarshi Das, Shehzaad Dhuliawala, Manzil Zaheer, Luke Vilnis, Ishan Durugkar,
  Akshay Krishnamurthy, Alex Smola, and Andrew McCallum. 2018.
\newblock Go for a walk and arrive at the answer: Reasoning over paths in
  knowledge bases using reinforcement learning.
\newblock In \emph{ICLR}.

\bibitem[{Dettmers et~al.(2018)Dettmers, Minervini, Stenetorp, and
  Riedel}]{dettmers2018convolutional}
Tim Dettmers, Pasquale Minervini, Pontus Stenetorp, and Sebastian Riedel. 2018.
\newblock Convolutional 2d knowledge graph embeddings.
\newblock In \emph{AAAI}.

\bibitem[{Ebisu and Ichise(2018)}]{ebisu2018toruse}
Takuma Ebisu and Ryutaro Ichise. 2018.
\newblock Toruse: Knowledge graph embedding on a lie group.
\newblock In \emph{AAAI}.

\bibitem[{Gal{\'a}rraga et~al.(2013)Gal{\'a}rraga, Teflioudi, Hose, and
  Suchanek}]{galarraga2013amie}
Luis~Antonio Gal{\'a}rraga, Christina Teflioudi, Katja Hose, and Fabian
  Suchanek. 2013.
\newblock Amie: association rule mining under incomplete evidence in
  ontological knowledge bases.
\newblock In \emph{WWW}.

\bibitem[{Garcia-Duran et~al.(2015)Garcia-Duran, Bordes, and
  Usunier}]{garcia2015composing}
Alberto Garcia-Duran, Antoine Bordes, and Nicolas Usunier. 2015.
\newblock Composing relationships with translations.
\newblock In \emph{EMNLP}.

\bibitem[{Glorot and Bengio(2010)}]{glorot2010understanding}
Xavier Glorot and Yoshua Bengio. 2010.
\newblock Understanding the difficulty of training deep feedforward neural
  networks.
\newblock In \emph{AISTATS}.

\bibitem[{Goodman(1961)}]{goodman1961snowball}
Leo~A Goodman. 1961.
\newblock Snowball sampling.
\newblock \emph{Ann. Math. Stat.}

\bibitem[{Guo et~al.(2019)Guo, Sun, and Hu}]{guo2019learning}
Lingbing Guo, Zequn Sun, and Wei Hu. 2019.
\newblock Learning to exploit long-term relational dependencies in knowledge
  graphs.
\newblock In \emph{ICML}.

\bibitem[{Guo et~al.(2015)Guo, Wang, Wang, Wang, and Guo}]{guo2015semantically}
Shu Guo, Quan Wang, Bin Wang, Lihong Wang, and Li~Guo. 2015.
\newblock Semantically smooth knowledge graph embedding.
\newblock In \emph{ACL-IJCNLP}.

\bibitem[{Guo et~al.(2018)Guo, Wang, Wang, Wang, and Guo}]{guo2018knowledge}
Shu Guo, Quan Wang, Lihong Wang, Bin Wang, and Li~Guo. 2018.
\newblock Knowledge graph embedding with iterative guidance from soft rules.
\newblock In \emph{AAAI}.

\bibitem[{Guu et~al.(2015)Guu, Miller, and Liang}]{guu2015traversing}
Kelvin Guu, John Miller, and Percy Liang. 2015.
\newblock Traversing knowledge graphs in vector space.
\newblock In \emph{EMNLP}.

\bibitem[{Hinton(1986)}]{hinton1986learning}
Geoffrey~E Hinton. 1986.
\newblock Learning distributed representations of concepts.
\newblock In \emph{CogSci}.

\bibitem[{Jain et~al.(2020)Jain, Rathi, Mausam, and
  Chakrabarti}]{jain2020knowledge}
Prachi Jain, Sushant Rathi, Mausam, and Soumen Chakrabarti. 2020.
\newblock Knowledge base completion: Baseline strikes back (again).
\newblock \emph{arXiv preprint arXiv:2005.00804}.

\bibitem[{Ji et~al.(2015)Ji, He, Xu, Liu, and Zhao}]{ji2015knowledge}
Guoliang Ji, Shizhu He, Liheng Xu, Kang Liu, and Jun Zhao. 2015.
\newblock Knowledge graph embedding via dynamic mapping matrix.
\newblock In \emph{ACL-IJCNLP}.

\bibitem[{Ji et~al.(2020)Ji, Pan, Cambria, Marttinen, and Yu}]{ji2020survey}
Shaoxiong Ji, Shirui Pan, Erik Cambria, Pekka Marttinen, and Philip~S Yu. 2020.
\newblock A survey on knowledge graphs: Representation, acquisition and
  applications.
\newblock \emph{arXiv preprint arXiv:2002.00388}.

\bibitem[{Jia et~al.(2016)Jia, Wang, Lin, Jin, and Cheng}]{jia2016locally}
Yantao Jia, Yuanzhuo Wang, Hailun Lin, Xiaolong Jin, and Xueqi Cheng. 2016.
\newblock Locally adaptive translation for knowledge graph embedding.
\newblock In \emph{AAAI}.

\bibitem[{Jiang et~al.(2019)Jiang, Wang, and Wang}]{jiang2019adaptive}
Xiaotian Jiang, Quan Wang, and Bin Wang. 2019.
\newblock Adaptive convolution for multi-relational learning.
\newblock In \emph{NAACL-HLT}.

\bibitem[{Kadlec et~al.(2017)Kadlec, Bajgar, and
  Kleindienst}]{kadlec2017knowledge}
Rudolf Kadlec, Ondrej Bajgar, and Jan Kleindienst. 2017.
\newblock Knowledge base completion: Baselines strike back.
\newblock In \emph{ACL RepL4NLP Workshop}.

\bibitem[{Kaffee et~al.(2017)Kaffee, Piscopo, Vougiouklis, Simperl, Carr, and
  Pintscher}]{kaffee2017glimpse}
Lucie-Aim{\'e}e Kaffee, Alessandro Piscopo, Pavlos Vougiouklis, Elena Simperl,
  Leslie Carr, and Lydia Pintscher. 2017.
\newblock A glimpse into babel: an analysis of multilinguality in wikidata.
\newblock In \emph{OpenSym}.

\bibitem[{Kazemi and Poole(2018)}]{kazemi2018simple}
Seyed~Mehran Kazemi and David Poole. 2018.
\newblock Simple embedding for link prediction in knowledge graphs.
\newblock In \emph{NeurIPS}.

\bibitem[{Kemp et~al.(2006)Kemp, Tenenbaum, Griffiths, Yamada, and
  Ueda}]{kemp2006learning}
Charles Kemp, Joshua~B Tenenbaum, Thomas~L Griffiths, Takeshi Yamada, and
  Naonori Ueda. 2006.
\newblock Learning systems of concepts with an infinite relational model.
\newblock In \emph{AAAI}.

\bibitem[{Lacroix et~al.(2018)Lacroix, Usunier, and
  Obozinski}]{lacroix2018canonical}
Timothee Lacroix, Nicolas Usunier, and Guillaume Obozinski. 2018.
\newblock Canonical tensor decomposition for knowledge base completion.
\newblock In \emph{ICML}.

\bibitem[{Lajus et~al.(2020)Lajus, Galárraga, and Suchanek}]{lajus2020fast}
Jonathan Lajus, Luis Galárraga, and Fabian~M. Suchanek. 2020.
\newblock Fast and exact rule mining with amie 3.
\newblock In \emph{ESWC}.

\bibitem[{Lin et~al.(2018)Lin, Socher, and Xiong}]{lin2018multi}
Xi~Victoria Lin, Richard Socher, and Caiming Xiong. 2018.
\newblock Multi-hop knowledge graph reasoning with reward shaping.
\newblock In \emph{EMNLP}.

\bibitem[{Lin et~al.(2015{\natexlab{a}})Lin, Liu, Luan, Sun, Rao, and
  Liu}]{lin2015modeling}
Yankai Lin, Zhiyuan Liu, Huanbo Luan, Maosong Sun, Siwei Rao, and Song Liu.
  2015{\natexlab{a}}.
\newblock Modeling relation paths for representation learning of knowledge
  bases.
\newblock In \emph{EMNLP}.

\bibitem[{Lin et~al.(2016)Lin, Liu, and Sun}]{lin2016knowledge}
Yankai Lin, Zhiyuan Liu, and Maosong Sun. 2016.
\newblock Knowledge representation learning with entities, attributes and
  relations.
\newblock In \emph{IJCAI}.

\bibitem[{Lin et~al.(2015{\natexlab{b}})Lin, Liu, Sun, Liu, and
  Zhu}]{lin2015learning}
Yankai Lin, Zhiyuan Liu, Maosong Sun, Yang Liu, and Xuan Zhu.
  2015{\natexlab{b}}.
\newblock Learning entity and relation embeddings for knowledge graph
  completion.
\newblock In \emph{AAAI}.

\bibitem[{Liu et~al.(2017)Liu, Wu, and Yang}]{liu2017analogical}
Hanxiao Liu, Yuexin Wu, and Yiming Yang. 2017.
\newblock Analogical inference for multi-relational embeddings.
\newblock In \emph{ICML}.

\bibitem[{Mahdisoltani et~al.(2014)Mahdisoltani, Biega, and
  Suchanek}]{mahdisoltani2014yago3}
Farzaneh Mahdisoltani, Joanna Biega, and Fabian Suchanek. 2014.
\newblock Yago3: A knowledge base from multilingual wikipedias.
\newblock In \emph{CIDR}.

\bibitem[{McCray(2003)}]{mccray2003upper}
Alexa~T McCray. 2003.
\newblock An upper-level ontology for the biomedical domain.
\newblock \emph{Comp. Funct. Genomics}, 4(1).

\bibitem[{Meilicke et~al.(2018)Meilicke, Fink, Wang, Ruffinelli, Gemulla, and
  Stuckenschmidt}]{meilicke2018fine}
Christian Meilicke, Manuel Fink, Yanjie Wang, Daniel Ruffinelli, Rainer
  Gemulla, and Heiner Stuckenschmidt. 2018.
\newblock Fine-grained evaluation of rule-and embedding-based systems for
  knowledge graph completion.
\newblock In \emph{ISWC}.

\bibitem[{Miller(1998)}]{miller1998wordnet}
George~A Miller. 1998.
\newblock \emph{WordNet: An electronic lexical database}.
\newblock MIT press.

\bibitem[{Mitchell et~al.(2018)Mitchell, Cohen, Hruschka, Talukdar, Yang,
  Betteridge, Carlson, Dalvi, Gardner, Kisiel et~al.}]{mitchell2018never}
Tom Mitchell, William Cohen, Estevam Hruschka, Partha Talukdar, Bishan Yang,
  Justin Betteridge, Andrew Carlson, Bhanava Dalvi, Matt Gardner, Bryan Kisiel,
  et~al. 2018.
\newblock Never-ending learning.
\newblock \emph{CACM}, 61(5):103--115.

\bibitem[{Nathani et~al.(2019)Nathani, Chauhan, Sharma, and
  Kaul}]{nathani2019learning}
Deepak Nathani, Jatin Chauhan, Charu Sharma, and Manohar Kaul. 2019.
\newblock Learning attention-based embeddings for relation prediction in
  knowledge graphs.
\newblock In \emph{ACL}.

\bibitem[{Nguyen et~al.(2016)Nguyen, Sirts, Qu, and
  Johnson}]{nguyen2016stranse}
Dat~Quoc Nguyen, Kairit Sirts, Lizhen Qu, and Mark Johnson. 2016.
\newblock Stranse: a novel embedding model of entities and relationships in
  knowledge bases.
\newblock In \emph{NAACL-HLT}.

\bibitem[{Nguyen et~al.(2018)Nguyen, Nguyen, Phung et~al.}]{nguyen2018novel}
Tu~Dinh Nguyen, Dat~Quoc Nguyen, Dinh Phung, et~al. 2018.
\newblock A novel embedding model for knowledge base completion based on
  convolutional neural network.
\newblock In \emph{NAACL-HLT}.

\bibitem[{Nickel et~al.(2015)Nickel, Murphy, Tresp, and
  Gabrilovich}]{nickel2015review}
Maximilian Nickel, Kevin Murphy, Volker Tresp, and Evgeniy Gabrilovich. 2015.
\newblock A review of relational machine learning for knowledge graphs.
\newblock \emph{Proc. IEEE}, 104(1).

\bibitem[{Nickel et~al.(2016)Nickel, Rosasco, and
  Poggio}]{nickel2016holographic}
Maximilian Nickel, Lorenzo Rosasco, and Tomaso Poggio. 2016.
\newblock Holographic embeddings of knowledge graphs.
\newblock In \emph{AAAI}.

\bibitem[{Nickel et~al.(2011)Nickel, Tresp, and Kriegel}]{nickel2011three}
Maximilian Nickel, Volker Tresp, and Hans-Peter Kriegel. 2011.
\newblock A three-way model for collective learning on multi-relational data.
\newblock In \emph{ICML}.

\bibitem[{Pezeshkpour et~al.(2020)Pezeshkpour, Tian, and
  Singh}]{pezeshkpour2020revisiting}
Pouya Pezeshkpour, Yifan Tian, and Sameer Singh. 2020.
\newblock Revisiting evaluation of knowledge base completion models.
\newblock In \emph{AKBC}.

\bibitem[{Ruffinelli et~al.(2020)Ruffinelli, Broscheit, and
  Gemulla}]{ruffinelli2020you}
Daniel Ruffinelli, Samuel Broscheit, and Rainer Gemulla. 2020.
\newblock You can teach an old dog new tricks! on training knowledge graph
  embeddings.
\newblock In \emph{ICLR}.

\bibitem[{Shi and Weninger(2017)}]{shi2017proje}
Baoxu Shi and Tim Weninger. 2017.
\newblock Proje: Embedding projection for knowledge graph completion.
\newblock In \emph{AAAI}.

\bibitem[{Socher et~al.(2013)Socher, Chen, Manning, and
  Ng}]{socher2013reasoning}
Richard Socher, Danqi Chen, Christopher~D Manning, and Andrew Ng. 2013.
\newblock Reasoning with neural tensor networks for knowledge base completion.
\newblock In \emph{NeurIPS}.

\bibitem[{Sun et~al.(2019)Sun, Deng, Nie, and Tang}]{sun2019rotate}
Zhiqing Sun, Zhi-Hong Deng, Jian-Yun Nie, and Jian Tang. 2019.
\newblock Rotate: Knowledge graph embedding by relational rotation in complex
  space.
\newblock In \emph{ICLR}.

\bibitem[{Tanon et~al.(2016)Tanon, Vrande{\v{c}}i{\'c}, Schaffert, Steiner, and
  Pintscher}]{tanon2016freebase}
Thomas~Pellissier Tanon, Denny Vrande{\v{c}}i{\'c}, Sebastian Schaffert, Thomas
  Steiner, and Lydia Pintscher. 2016.
\newblock From freebase to wikidata: The great migration.
\newblock In \emph{WWW}.

\bibitem[{Toutanova and Chen(2015)}]{toutanova2015observed}
Kristina Toutanova and Danqi Chen. 2015.
\newblock Observed versus latent features for knowledge base and text
  inference.
\newblock In \emph{ACL CVSC Workshop}.

\bibitem[{Toutanova et~al.(2015)Toutanova, Chen, Pantel, Poon, Choudhury, and
  Gamon}]{toutanova2015representing}
Kristina Toutanova, Danqi Chen, Patrick Pantel, Hoifung Poon, Pallavi
  Choudhury, and Michael Gamon. 2015.
\newblock Representing text for joint embedding of text and knowledge bases.
\newblock In \emph{EMNLP}.

\bibitem[{Trouillon et~al.(2019)Trouillon, Gaussier, Dance, and
  Bouchard}]{trouillon2019inductive}
Th{\'e}o Trouillon, {\'E}ric Gaussier, Christopher~R Dance, and Guillaume
  Bouchard. 2019.
\newblock On inductive abilities of latent factor models for relational
  learning.
\newblock \emph{JAIR}, 64.

\bibitem[{Trouillon et~al.(2016)Trouillon, Welbl, Riedel, Gaussier, and
  Bouchard}]{trouillon2016complex}
Th{\'e}o Trouillon, Johannes Welbl, Sebastian Riedel, {\'E}ric Gaussier, and
  Guillaume Bouchard. 2016.
\newblock Complex embeddings for simple link prediction.
\newblock In \emph{ICML}.

\bibitem[{Vashishth et~al.(2020{\natexlab{a}})Vashishth, Sanyal, Nitin,
  Agrawal, and Talukdar}]{vashishth2020interacte}
Shikhar Vashishth, Soumya Sanyal, Vikram Nitin, Nilesh Agrawal, and Partha
  Talukdar. 2020{\natexlab{a}}.
\newblock Interacte: Improving convolution-based knowledge graph embeddings by
  increasing feature interactions.
\newblock In \emph{AAAI}.

\bibitem[{Vashishth et~al.(2020{\natexlab{b}})Vashishth, Sanyal, Nitin, and
  Talukdar}]{vashishth2020composition}
Shikhar Vashishth, Soumya Sanyal, Vikram Nitin, and Partha Talukdar.
  2020{\natexlab{b}}.
\newblock Composition-based multi-relational graph convolutional networks.
\newblock In \emph{ICLR}.

\bibitem[{Vrande{\v{c}}i{\'c} and Kr{\"o}tzsch(2014)}]{vrandevcic2014wikidata}
Denny Vrande{\v{c}}i{\'c} and Markus Kr{\"o}tzsch. 2014.
\newblock Wikidata: a free collaborative knowledgebase.
\newblock \emph{CACM}, 57(10).

\bibitem[{Vu et~al.(2019)Vu, Nguyen, Nguyen, Phung et~al.}]{vu2019capsule}
Thanh Vu, Tu~Dinh Nguyen, Dat~Quoc Nguyen, Dinh Phung, et~al. 2019.
\newblock A capsule network-based embedding model for knowledge graph
  completion and search personalization.
\newblock In \emph{NAACL-HLT}.

\bibitem[{Wang et~al.(2017)Wang, Mao, Wang, and Guo}]{wang2017knowledge}
Quan Wang, Zhendong Mao, Bin Wang, and Li~Guo. 2017.
\newblock Knowledge graph embedding: A survey of approaches and applications.
\newblock \emph{TKDE}, 29(12).

\bibitem[{Wang et~al.(2015)Wang, Wang, and Guo}]{wang2015knowledge}
Quan Wang, Bin Wang, and Li~Guo. 2015.
\newblock Knowledge base completion using embeddings and rules.
\newblock In \emph{IJCAI}.

\bibitem[{Wang and Cohen(2016)}]{wang2016learning}
William~Yang Wang and William~W Cohen. 2016.
\newblock Learning first-order logic embeddings via matrix factorization.
\newblock In \emph{IJCAI}.

\bibitem[{Wang et~al.(2014)Wang, Zhang, Feng, and Chen}]{wang2014knowledge}
Zhen Wang, Jianwen Zhang, Jianlin Feng, and Zheng Chen. 2014.
\newblock Knowledge graph embedding by translating on hyperplanes.
\newblock In \emph{AAAI}.

\bibitem[{Xiao et~al.(2016{\natexlab{a}})Xiao, Huang, and Zhu}]{xiao2016one}
Han Xiao, Minlie Huang, and Xiaoyan Zhu. 2016{\natexlab{a}}.
\newblock From one point to a manifold: knowledge graph embedding for precise
  link prediction.
\newblock In \emph{IJCAI}.

\bibitem[{Xiao et~al.(2016{\natexlab{b}})Xiao, Huang, and Zhu}]{xiao2016transg}
Han Xiao, Minlie Huang, and Xiaoyan Zhu. 2016{\natexlab{b}}.
\newblock Transg: A generative model for knowledge graph embedding.
\newblock In \emph{ACL}.

\bibitem[{Xie et~al.(2016)Xie, Liu, and Sun}]{xie2016representation}
Ruobing Xie, Zhiyuan Liu, and Maosong Sun. 2016.
\newblock Representation learning of knowledge graphs with hierarchical types.
\newblock In \emph{IJCAI}.

\bibitem[{Xiong et~al.(2017)Xiong, Hoang, and Wang}]{xiong2017deeppath}
Wenhan Xiong, Thien Hoang, and William~Yang Wang. 2017.
\newblock Deeppath: A reinforcement learning method for knowledge graph
  reasoning.
\newblock In \emph{EMNLP}.

\bibitem[{Xu and Li(2019)}]{xu2019relation}
Canran Xu and Ruijiang Li. 2019.
\newblock Relation embedding with dihedral group in knowledge graph.
\newblock In \emph{ACL}.

\bibitem[{Yang et~al.(2015)Yang, Yih, He, Gao, and Deng}]{yang2014embedding}
Bishan Yang, Wen-tau Yih, Xiaodong He, Jianfeng Gao, and Li~Deng. 2015.
\newblock Embedding entities and relations for learning and inference in
  knowledge bases.
\newblock In \emph{ICLR}.

\bibitem[{Zhang et~al.(2019)Zhang, Tay, Yao, and Liu}]{zhang2019quaternion}
Shuai Zhang, Yi~Tay, Lina Yao, and Qi~Liu. 2019.
\newblock Quaternion knowledge graph embeddings.
\newblock In \emph{NeurIPS}.

\bibitem[{Zhang et~al.(2020)Zhang, Zhuang, Zhu, Shi, Xiong, and
  He}]{zhang2020relational}
Zhao Zhang, Fuzhen Zhuang, Hengshu Zhu, Zhiping Shi, Hui Xiong, and Qing He.
  2020.
\newblock Relational graph neural network with hierarchical attention for
  knowledge graph completion.
\newblock In \emph{AAAI}.

\end{thebibliography}
\bibliographystyle{acl_natbib}

\clearpage

\appendix
\section{Literature review}
\label{appx:related}

Table~\ref{table:lit-review} provides an overview of knowledge graph embedding papers with respect to datasets and evaluation tasks.
In our review, we only consider papers published between 2014 and 2020 in the main proceedings of conferences where \task{} embedding papers are most likely to appear: Artificial intelligence (AAAI, IJCAI), machine learning (ICML, ICLR, NeurIPS), and natural language processing (ACL, EMNLP, NAACL).

The main evaluation benchmarks are 
\textbf{\fb{}} \cite{bordes2013translating},
\textbf{\wn{}} \cite{bordes2013translating},
\textbf{\fb{}-237} \cite{toutanova2015observed},
\textbf{\wn{}RR} \cite{dettmers2018convolutional},
\textbf{FB13} \cite{socher2013reasoning},
\textbf{WN11} \cite{socher2013reasoning},
\textbf{NELL-995} \cite{xiong2017deeppath},
\textbf{YAGO3-10} \cite{dettmers2018convolutional},
\textbf{Countries} \cite{bouchard2015approximate}.
\textbf{UMLS} \cite{mccray2003upper},
\textbf{Kinship} \cite{kemp2006learning},
\textbf{Families} \cite{hinton1986learning}, and 
other versions of \textbf{NELL}~\cite{mitchell2018never}.

\newcolumntype{V}{>{\centering\arraybackslash}p{3.25cm}}

\begin{table*}
\centering
\caption{
An overview of knowledge graph embedding papers published between 2014 and 2020 with respect to datasets and evaluation tasks.
Original citations for datasets are given in Appendix~\ref{appx:related}.
Link pred. refers to link prediction, and triple class. refers to triple classification, both of which are covered in \S~\ref{sec:benchmarking}.
}
\label{table:lit-review}
\resizebox{\textwidth}{!}{
\def\arraystretch{1.2}
\begin{tabular}{ p{0.3cm} p{4.5cm} cccccc>{\centering}p{5.5cm}c 
>{\centering}p{0.7cm}>{\centering}p{0.7cm} V}  
\toprule
 & \multirow{9}{*}{\textbf{Reference}} &  \multicolumn{7}{c}{\textbf{Datasets}}   && \multicolumn{3}{c}{\textbf{Evaluation tasks}} \\ 
 \cline{3-9} \cline{11-13}
     & &  \rot{90}{FB15K} & \rot{90}{FB15K-237\,\,\,} & \rot{90}{FB13} & \rot{90}{WN18} & \rot{90}{WN18RR} & \rot{90}{WN11} & Other   && \rot{90}{Link pred.} & \rot{90}{Triple class.} & Other \\
    \toprule
    \multirow{17}{*}{\rotatebox[origin=c]{90}{\textbf{AAAI, IJCAI}}}
    & \multirow{2}{*}{\cite{wang2014knowledge}} & \multirow{2}{*}{\cmark} &  & \multirow{2}{*}{\cmark} & \multirow{2}{*}{\cmark} &  & \multirow{2}{*}{\cmark} & \multirow{2}{*}{FB5M} && \multirow{2}{*}{\cmark} & \multirow{2}{*}{\cmark} & relation extraction (FB5M) \\ 
    \cline{3-9} \cline{11-13}
    & \multirow{2}{*}{\cite{lin2015learning}} & \multirow{2}{*}{\cmark} &  & \multirow{2}{*}{\cmark} & \multirow{2}{*}{\cmark} &  & \multirow{2}{*}{\cmark} &  \multirow{2}{*}{FB40K} && \multirow{2}{*}{\cmark} & \multirow{2}{*}{\cmark} & relation extraction (FB40K) \\
    \cline{3-9} \cline{11-13}
    & \cite{wang2015knowledge} &  &  &  &  &  &  &  NELL (Location, Sports) && \cmark &  &  \\
    \cline{3-9} \cline{11-13}
    \cline{3-9} \cline{11-13}
    & \cite{nickel2016holographic} & \cmark &  &  & \cmark &  &  & Countries && \cmark &   & \\
    \cline{3-9} \cline{11-13}
    & \cite{lin2016knowledge} &  &  &  &  &  &  &  FB24K && \cmark &   &  \\
    \cline{3-9} \cline{11-13}
    & \cite{wang2016learning} &  \cmark &  &  & \cmark &  &  &   && \cmark &   &  \\
    \cline{3-9} \cline{11-13}
    & \cite{xiao2016one} &  \cmark &  & \cmark & \cmark & & \cmark  &  && \cmark & \cmark  &  \\
    \cline{3-9} \cline{11-13}
    & \cite{jia2016locally} & \cmark &  & \cmark & \cmark &  & \cmark &  && \cmark & \cmark &  \\
    \cline{3-9} \cline{11-13}
    & \cite{xie2016representation} & \cmark & & & & & & \fb+ && \cmark & \cmark & \\
    \cline{3-9} \cline{11-13}
    & \multirow{2}{*}{\cite{shi2017proje}} & \multirow{2}{*}{\cmark} &  &  &  &  &  &  \multirow{2}{*}{SemMedDB, DBPedia} && \multirow{2}{*}{\cmark} &  & fact checking (not on \fb{}) \\
    \cline{3-9} \cline{11-13}
    & \cite{dettmers2018convolutional} &  \cmark & \cmark &  & \cmark & \cmark & &  YAGO3-10, Countries && \cmark &   &  \\
    \cline{3-9} \cline{11-13}
    & \cite{ebisu2018toruse} &   \cmark &  &  & \cmark &  &  &   && \cmark &   &  \\
    \cline{3-9} \cline{11-13}
    & \cite{guo2018knowledge} &   \cmark &  &  &  &  &  & YAGO37 && \cmark &   &  \\
    \cline{3-9} \cline{11-13}
    & \cite{zhang2020relational} &   \cmark & \cmark &  & \cmark & \cmark &  &   && \cmark &   &  \\
    \cline{3-9} \cline{11-13}
    & \cite{vashishth2020interacte} &   & \cmark &  &  & \cmark &  &  YAGO3-10 && \cmark &   &  \\ 
    \toprule
    \multirow{14}{*}{\rotatebox[origin=c]{90}{\textbf{ICML, ICLR, NeurIPS}}}
    & \multirow{2}{*}{\cite{yang2014embedding}} &  \multirow{2}{*}{\cmark} &  &  & \multirow{2}{*}{\cmark} &  &  &  \multirow{2}{*}{FB15K-401} && \multirow{2}{*}{\cmark} &   & rule extraction (FB15K-401) \\ 
    \cline{3-9} \cline{11-13}
    & \cite{trouillon2016complex} &  \cmark &  &  & \cmark &  &  &   && \cmark &   &   \\
    \cline{3-9} \cline{11-13}
    & \cite{liu2017analogical} &  \cmark &  &  & \cmark &  &  &   && \cmark &   &  \\
    \cline{3-9} \cline{11-13}
    & \cite{kazemi2018simple} & \cmark &  &  & \cmark &  &  &   && \cmark &   &  \\
    \cline{3-9} \cline{11-13}
    & \multirow{2}{*}{\cite{das2017go}} &  & \multirow{2}{*}{\cmark} &  &  & \multirow{2}{*}{\cmark} &  &  NELL-995, UMLS, Kinship, Countries, WikiMovies && \multirow{2}{*}{\cmark} 
    &   & \multirow{2}{*}{QA (WikiMovies)}  \\
    \cline{3-9} \cline{11-13}
    & \cite{lacroix2018canonical} & \cmark & \cmark &  & \cmark & \cmark &  &  YAGO3-10 && \cmark &   &  \\
    \cline{3-9} \cline{11-13}
    & \multirow{2}{*}{\cite{guo2019learning}} &  \multirow{2}{*}{\cmark} & \multirow{2}{*}{\cmark} &  & \multirow{2}{*}{\cmark} &  &  &  DBPedia-YAGO3, DBPedia-Wikidata && \multirow{2}{*}{\cmark} &   & entity alignment (DBPedia graphs) \\
    \cline{3-9} \cline{11-13}
    & \cite{sun2019rotate} & \cmark & \cmark &  & \cmark & \cmark &  &   && \cmark &   &  \\
    \cline{3-9} \cline{11-13}
    & \cite{zhang2019quaternion} &  \cmark & \cmark &  & \cmark & \cmark &  &   && \cmark &   &  \\
    \cline{3-9} \cline{11-13}
    & \cite{balazevic2019multi} & & \cmark{} & & &\cmark{} & && & \cmark{} & & \\ 
    \cline{3-9} \cline{11-13}
    & \multirow{2}{*}{\cite{vashishth2020composition}} &   & \multirow{2}{*}{\cmark} &  &  & \multirow{2}{*}{\cmark}  &  &  \multirow{2}{*}{MUTAG, AM, PTC} && \multirow{2}{*}{\cmark} &   & graph classification (MUTAG, AM, PTC) \\ 
    \toprule
    \multirow{17}{*}{\rotatebox[origin=c]{90}{\textbf{ACL, EMNLP, NAACL}}}
    & \cite{ji2015knowledge} &  \cmark &  & \cmark & \cmark &  & \cmark &  && \cmark & \cmark &  \\
    \cline{3-9} \cline{11-13}
    & \cite{guo2015semantically} &   &  &  &  &  &  &  NELL (Location, Sports, Freq) && \cmark & \cmark &  \\
    \cline{3-9} \cline{11-13}
    & \cite{guu2015traversing} &  &  & \cmark &  &  & \cmark &   && \cmark & \cmark  &  \\
    \cline{3-9} \cline{11-13}
    & \cite{garcia2015composing} & \cmark &  &  &  &  &  & Families && \cmark &   &  \\
    \cline{3-9} \cline{11-13}
    & \multirow{2}{*}{\cite{lin2015modeling}} &  \multirow{2}{*}{\cmark} &  &  &  &  &   & \multirow{2}{*}{FB40K} && \multirow{2}{*}{\cmark} &   & relation extraction (FB40K)   \\
    \cline{3-9} \cline{11-13}
    & \cite{xiao2016transg} &  \cmark &  & \cmark & \cmark &  & \cmark &  && \cmark &  \cmark  &  \\
    \cline{3-9} \cline{11-13}
    & \cite{nguyen2016stranse} &  \cmark &  &  & \cmark &  &  &  && \cmark &   &  \\
    \cline{3-9} \cline{11-13}
    & \cite{xiong2017deeppath} &  & \cmark &  &  &  &  & NELL-995  && \cmark &   & rule mining   \\
    \cline{3-9} \cline{11-13}
    & \cite{lin2018multi} &  & \cmark &  &  & \cmark  & &  NELL-995, UMLS, Kinship && \cmark &   &   \\
    \cline{3-9} \cline{11-13}
    & \cite{nguyen2018novel} &  & \cmark &  &  & \cmark & &   && \cmark &   &   \\
    \cline{3-9} \cline{11-13}
    & \cite{bansal2019a2n} &   & \cmark &  &  & \cmark &  &   && \cmark &   &  \\
    \cline{3-9} \cline{11-13}
    & \cite{xu2019relation} &  \cmark & \cmark &  & \cmark & \cmark &  &  YAGO3-10, Family && \cmark &   &   \\
    \cline{3-9} \cline{11-13}
    & \cite{balazevic2019tucker} &  \cmark & \cmark &  & \cmark & \cmark &  &   && \cmark &   &   \\
    \cline{3-9} \cline{11-13}
    & \multirow{2}{*}{\cite{vu2019capsule}} &  & \multirow{2}{*}{\cmark} &  &  & \multirow{2}{*}{\cmark} &  &  \multirow{2}{*}{SEARCH17} && \multirow{2}{*}{\cmark} &   & personalized search (SEARCH17)  \\
    \cline{3-9} \cline{11-13}
    & \cite{nathani2019learning} &   & \cmark &  &  & \cmark  & &  NELL-995, UMLS, Kinship && \cmark &   &   \\
    \cline{3-9} \cline{11-13}
    & \cite{jiang2019adaptive} &  \cmark & \cmark &  & \cmark & \cmark  & &   && \cmark &   &   \\
 \bottomrule
\end{tabular}
}
\end{table*}

\section{Seeds for data collection}
\label{appx:seeds}

Table~\ref{table:seeds} provides all seed entity and relation types used to collect \benchmark{}. 
Each type is given first by its natural language label and then by its Wikidata unique ID: Entity IDs begin with Q, whereas relation (property) IDs begin with P. 
For the entity types that apply to \emph{people} (e.g., actor, musician, journalist), we retrieved seed entities by querying Wikidata using the \emph{occupation} relation.
For the entity types that apply to \emph{things} (e.g., airline, disease, tourist attraction), we retrieved seed entities by querying Wikidata using the \emph{instance of} and \emph{subclass of} relations. 

\begin{table*}
\centering
\caption{The entity and relation types (Wikidata IDs in parentheses) used to seed \benchmark{}. 
}
\label{table:seeds}
\resizebox{\textwidth}{!}{
\def\arraystretch{1.1}
\begin{tabular}{ p{0.5cm} p{17cm} } 
\toprule
  &  {Seed types} \\
\toprule
    \multirow{6}{*}{\rotatebox[origin=c]{90}{{Entities}}}
    &  {actor} (Q33999), {airline} (Q46970), {airport} (Q1248784), {athlete} (Q2066131), {book} (Q571), {businessperson} (Q43845), {city} (Q515), {company} (Q783794), {country} (Q6256), {disease} (Q12136), {engineer} (Q81096), {film} (Q11424), {government agency} (Q327333), {journalist} (Q1930187), {lake} (Q23397), {monarch} (Q116), {mountain} (Q8502), {musical group} (Q215380), {musician} (Q639669), {newspaper} (Q11032), {ocean} (Q9430), {politician} (Q82955), {record label} (Q18127), {religion} (Q9174), {religious leader} (Q15995642), {religious text} (Q179461), {scientist} (Q901), {sports league} (Q623109), {sports team} (Q12973014), {stadium} (Q483110), {television program} (Q15416), {tourist attraction} (Q570116), {visual artist} (Q3391743), {visual artwork} (Q4502142), {writer} (Q36180) \\ 
    \midrule
    \multirow{14}{*}{\rotatebox[origin=c]{90}{{Relations}}} & {airline alliance} (P114), {airline hub} (P113), {architect} (P84), {architectural style} (P149), {author} (P50), {capital} (P36), {cast member} (P161), {cause of death} (P509), {chairperson} (P488), {chief executive officer} (P169), {child} (P40), {continent} (P30), {country} (P17), {country of citizenship} (P27), {country of origin} (P495), {creator} (P170), {diplomatic relation} (P530), {director} (P57), {drug used for treatment} (P2176), {educated at} (P69), {employer} (P108), {ethnic group} (P172), {field of work} (P101), {foundational text} (P457), {founded by} (P112), {genre} (P136), {head of government} (P6), {head of state} (P35), {headquarters location} (P159), {health specialty} (P1995), {indigenous to} (P2341), {industry} (P452), {influenced by} (P737), {instance of} (P31), {instrument} (P1303), {language of work or name} (P407), {languages spoken, written, or signed} (P1412), {legal form} (P1454), {legislative body} (P194), {located in the administrative terroritorial entity} (P131), {location of formation} (P740), {medical condition} (P1050), {medical examinations} (P923), {member of} (P463), {member of political party} (P102), {member of sports team} (P54), {mountain range} (P4552), {movement} (P135), {named after} (P138), {narrative location} (P840), {notable works} (P800), {occupant} (P466), {occupation} (P106), {official language} (P37), {parent organization} (P749), {part of} (P361), {place of birth} (P19), {place of burial} (P119), {place of death} (P20), {practiced by} (P3095), {product or material produced} (P1056), {publisher} (P123), {record label} (P264), {regulated by} (P3719), {religion} (P140), {residence} (P551), {shares border with} (P47), {sibling} (P3373), {sport} (P641), {spouse} (P26), {studies} (P2578), {subclass of} (P279), {symptoms} (P780), {time period} (P2348), {tributary} (P974), {unmarried partner} (P451), {use} (P366), {uses} (P2283) \\ 
 \bottomrule
\end{tabular}
}
\end{table*}

\section{Negative annotation guidelines}
\label{appx:guidelines}

We provide the annotation guidelines we used to label candidate negative triples (\S~\ref{sec:negative-data-collection}).

\paragraph{Task}
You must label each triple as either \emph{true} or \emph{false}. To help you find the answer, we have provided you with Wikipedia and Wikidata links for the entities and relations in each triple. You may also search on Google for the answer, although most claims should be resolvable using Wikipedia and Wikidata alone. If you are not able to find any reliable, specific, clear information supporting the claim, choose \emph{false}.
You may explain your reasoning if need be or provide sources to back up your answer in the optional explanation column. 

\paragraph{Examples}
False triples may have problems with grammar, factual content, or both. 
Examples of \emph{grammatically incorrect} triples are those whose entity or relation types do not make sense, for example:
\begin{itemize}
    \item (\emph{United States of America}, \emph{continent}, \emph{science fiction writer})
    \item (\emph{Mohandas Karamchand Gandhi}, \emph{medical condition}, \emph{British Raj})
    \item (\emph{Canada}, \emph{foundational text}, \emph{Vietnamese cuisine})
\end{itemize}
Examples of grammatically correct but \emph{factually false} triples include:
\begin{itemize}
    \item (\emph{United States of America}, \emph{continent}, \emph{Europe})
    \item (\emph{Mohandas Karamchand Gandhi}, \emph{country of citizenship}, \emph{Argentina})
    \item (\emph{Canada}, \emph{foundational text}, \emph{Harry Potter and the Goblet of Fire})
    \item (\emph{Alexander Pushkin}, \emph{influenced by}, \emph{Leo Tolstoy}) --- Pushkin died only a few years after Tolstoy was born, so this sentence is unlikely.
\end{itemize}
Notice that in the latter examples, the entity types match up, but the statements are still false. 

\paragraph{Tips}
For triples about people's \emph{occupation} and \emph{genre}, try to be as specific as possible. For example, if the triple says (<person>, \emph{occupation}, \emph{guitarist}) but that person is mainly known for their singing, choose \emph{false}, even if that person plays the guitar. 
Likewise, if a triple says (<person>, \emph{genre}, \emph{classical}) but they are mostly known for jazz music, choose \emph{false} even if, for example, that person had classical training in their childhood.

\section{Embedding models}
\label{appx:models}

We briefly overview the \update{five} models compared in our link prediction and triple classification tasks. 

\newpar{RESCAL} \cite{nickel2011three} was one of the first knowledge graph embedding models.
Although it is not often used as a baseline, ~\citet{ruffinelli2020you} showed that it is competitive when appropriately tuned. 
RESCAL treats relational learning as tensor decomposition, scoring entity embeddings $\mathbf{h},\mathbf{r} \in \mathbb{R}^{d_e}$ and relation embeddings $\mathbf{R} \in \mathbb{R}^{d_e \times d_e}$ with the bilinear form $\mathbf{h}^{\top}\mathbf{R}\mathbf{t}$.

\newpar{TransE} \cite{bordes2013translating}
treats relations as translations between entities, i.e., $\mathbf{h} + \mathbf{r} \approx \mathbf{t}$ for $\mathbf{h},\mathbf{r}, \mathbf{t} \in \mathbb{R}^{d_e}$, and scores embeddings with negative Euclidean distance $-\| \mathbf{h} + \mathbf{r} - \mathbf{t}\|$. 
TransE is likely the most popular baseline for \task{} tasks and the most influential of all \task{} embedding papers. 

\newpar{ComplEx} \cite{trouillon2016complex} uses a bilinear function to score triples with a diagonal relation embedding matrix and complex-valued embeddings. 
Its scoring function is  $\text{re}\left(\mathbf{h}^{\top} \text{diag}(\mathbf{r}) \mathbf{\overline{t}}\right)$, where $\mathbf{\overline{t}}$ is the complex conjugate of $\mathbf{t}$ and re denotes the real part of a complex number.

\newpar{ConvE} \cite{dettmers2018convolutional} is one of the first and most popular \emph{nonlinear} models for \task{}.
It concatenates head and relation embeddings $\mathbf{h}$ and $\mathbf{r}$ into a two-dimensional ``image'', applies a pointwise linearity over convolutional and fully-connected layers, and multiplies the result with the tail embedding $\mathbf{t}$ to obtain a score. 
\update{Formally, its scoring function is given as $f(\textrm{vec} (f([\overline{\mathbf{h}}; \overline{\mathbf{r}}] * \omega))\mathbf{W})\mathbf{t}$, where $f$ is a nonlinearity (originally, ReLU), $[\overline{\mathbf{h}}; \overline{\mathbf{r}}]$ denotes a concatenation and two-dimensional reshaping of the head and relation embeddings, $\omega$ denotes the filters of the convolutional layer, and $\textrm{vec}$ denotes the flattening of a two-dimensional matrix.}

\newpar{TuckER} \cite{balazevic2019tucker} \update{is a linear model based on the Tucker tensor decomposition, which factorizes a tensor into three lower-rank matrices and a core tensor.
The TuckER scoring function for a single triple $(h, r, t)$ is given as $\mathcal{W} \times_1 \mathbf{h} \times_2 \mathbf{r} \times_3 \mathbf{t}$, where $\mathcal{W}$ is the mode-three core tensor that is shared among all entity and relation embeddings, and $\times_n$ denotes the tensor product along the $n$-th mode of the tensor. 
TuckER can be seen as a generalized form of other linear \task{} embedding models like RESCAL and ComplEx.
}

\section{Content comparison}
\label{appx:lp-comparison}

We provide additional comparison of the contents in \benchmarkMed{} and \fb-237.

\begin{figure*}[t!]
    \centering
    \includegraphics[width=\textwidth]{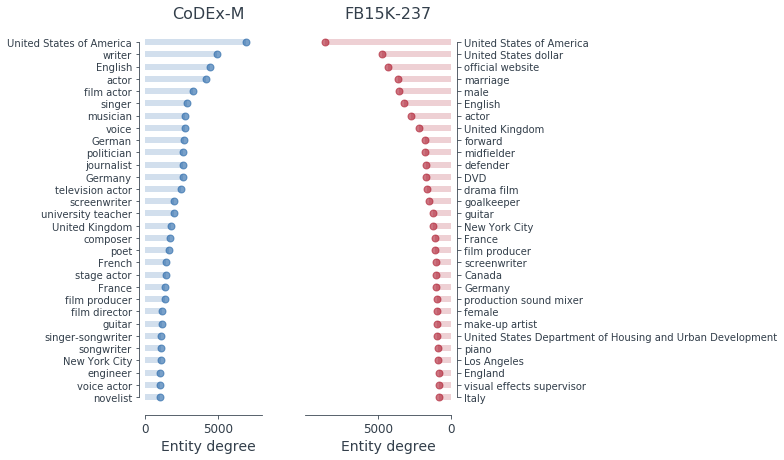}
    \caption{Top-30 \textbf{most frequent entities} in \benchmarkMed{} and \fb-237.
    }
    \label{fig:top-k-entities}
\end{figure*}

\begin{figure*}[t!]
    \centering
    \includegraphics[width=\textwidth]{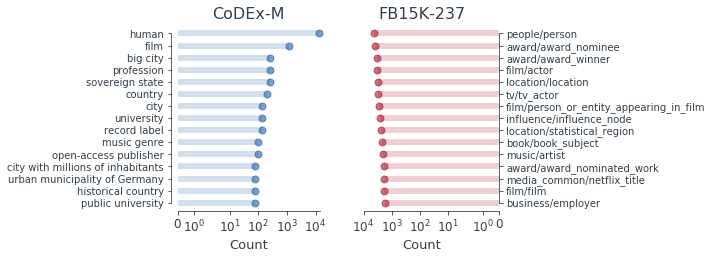}
    \caption{Top-15 \textbf{most frequent entity types} in \benchmarkMed{} and \fb-237.
    }
    \label{fig:top-k-entity-types}
\end{figure*}

Figure~\ref{fig:top-k-entities}, which plots the \textbf{top-30 entities by frequency} in the two benchmarks, demonstrates that both dataset are biased toward developed Western countries and cultures.
However, \benchmarkMed{} is more diverse in domain.
It covers academia, entertainment, journalism, politics, science, and writing, whereas \fb-237 covers mostly entertaiment and sports. 
\fb-237 is also much more biased toward the United States in particular, as five of its top-30 entities are specific to the US: \emph{United States of America}, \emph{United States dollar}, \emph{New York City}, \emph{Los Angeles}, and the \emph{United States Department of Housing and Urban Development}.

Figure~\ref{fig:top-k-entity-types} compares the \textbf{top-15 entity types} in \benchmarkMed{} and \fb-237.
Again, \benchmarkMed{} is diverse, covering people, places, organizations, movies, and abstract concepts, whereas \fb-237 has many overlapping entity types mostly about entertainment.

\section{Hyperparameter search}
\label{appx:hyperparameters}

Table~\ref{table:hyperparameter-search-space} gives our hyperparameter search space. 
Tables~\ref{table:lp-small}, \ref{table:lp-medium}, and  \ref{table:lp-large} report the best hyperparameter configurations for link prediction on \benchmarkSmall{}, \benchmarkMed{}, and \benchmarkLarge{}, respectively.
Tables~\ref{table:tc-small} and \ref{table:tc-medium} report the best hyperparameter configurations for triple classification on the hard negatives in \benchmarkSmall{} and \benchmarkMed{}, respectively.

\paragraph{Terminology}
For embedding initialization, \textbf{Xv} refers to Xavier initialization~\cite{glorot2010understanding}. 
The \textbf{reciprocal relations} model refers to learning separate relation embeddings for queries in the direction of $(h, r, ?)$ versus $(?, r, t)$~\cite{kazemi2018simple}. 
The \textbf{frequency weighting} regularization technique refers to regularizing embeddings by the relative frequency of the corresponding entity or relation in the training data. 

\paragraph{Search strategies}
Recall that we select models using Ax, which supports hyperparameter search using both quasi-random sequences of generated configurations and Bayesian optimization (BO). 
The search strategy for each \benchmark{} dataset is as follows: 
\begin{itemize}
    \item \textbf{\benchmarkSmall{}}: Per negative sampling type/loss combination, we generate 30 quasi-random trials followed by 10 BO trials.
    We select the best-performing model by validation MRR over all such combinations. 
    In each trial, the model is trained for a maximum of 400 epochs with an early stopping patience of 5. 
    We also terminate a trial after 50 epochs if the model does not reach $\geq$ 0.05 MRR.
    \item \textbf{\benchmarkMed{}}: Per negative sampling type/loss combination, we generate 20 quasi-random trials.
    The maximum number of epochs and early stopping criteria are the same as for \benchmarkSmall{}. 
    \item \textbf{\benchmarkLarge{}}: Per negative sampling type/loss combination, we generate 10 quasi-random trials of 20 training epochs instead of 400.
    We reduce the number of epochs to limit resource usage.
    In most cases, MRR plateaus after 20-30 epochs, an observation which is consistent with~\cite{ruffinelli2020you}. 
    Then, we take the best-performing model by validation MRR over all such combinations, and retrain that model for a maximum of 400 epochs. 
\end{itemize}
Note that we search using MRR as our metric, but the triple classification task measures 0/1 accuracy, not ranking performance.
For triple classification, we choose the model with the highest validation accuracy among the pre-trained models across all negative sampling type/loss function combinations. 

We release all pretrained LibKGE models and accompanying configuration files in the centralized \benchmark{} repository. 

\begin{table*}[t!]
\centering
\caption{Our hyperparameter search space.
We follow the naming conventions and ranges given by~\citet{ruffinelli2020you}, and explain the meanings of selected hyperparameter settings in Appendix~\ref{appx:hyperparameters}. 
As most \task{} embedding models have a wide range of configuration options, we encourage future work to follow this tabular scheme for transparent reporting of implementation details. 
}
\label{table:hyperparameter-search-space}
\resizebox{0.7\textwidth}{!}{
\begin{tabular}{ l r } 
\toprule
Hyperparameter & Range \\ 
\toprule
Embedding size & \{$128, 256, 512$\} \\ 
Training type & \{NegSamp, 1vsAll, KvsAll\} \\ 
\quad Reciprocal & \{True, False\}  \\ 
\quad \# head samples (NegSamp) & $[1, 1000]$, log scale \\
\quad \# tail samples (NegSamp) & $[1, 1000]$, log scale \\
\quad Label smoothing (KvsAll) & $[0, 0.3]$ \\
Loss & \{MR, BCE, CE\} \\ 
\quad Margin (MR) & $[0, 10]$ \\ 
\quad $\ell_p$ norm (TransE) & \{$1, 2$\} \\ 
Optimizer & \{Adam, Adagrad\} \\
\quad Batch size & \{$128, 256, 512, 1024$\} \\ 
\quad Learning rate & $[10^{-4}, 1]$, log scale \\
\quad LR scheduler patience & $[0, 10]$ \\ 
$\ell_p$ regularization & \{$1, 2, 3,$ None\} \\ 
\quad Entity embedding weight & $[10^{20}, 10^{-5}]$ \\ 
\quad Relation embedding weight & $[10^{20}, 10^{-5}]$ \\ 
\quad Frequency weighting & \{True, False\} \\ 
Embedding normalization (TransE) & \\ 
\quad Entity & \{True, False\}  \\ 
\quad Relation & \{True, False\} \\
Dropout & \\ 
\quad Entity embedding & $[0.0, 0.5]$  \\
\quad Relation embedding & $[0.0, 0.5]$  \\
\quad Feature map (ConvE) & $[0.0, 0.5]$  \\ 
\quad Projection (ConvE) & $[0.0, 0.5]$ \\ 
Embedding initialization & \{Normal, Unif, XvNorm, XvUnif\}  \\ 
\quad Stdev (Normal) & $[10^{-5}, 1.0]$  \\ 
\quad Interval (Unif) & $[-1.0, 1.0]$ \\ 
\quad Gain (XvNorm) & $1.0$ \\ 
\quad Gain (XvUnif) & $1.0$ \\ 
\bottomrule
\end{tabular}

}
\end{table*}

\begin{table*}[t!]
\centering
\caption{Best \textbf{link prediction} hyperparameter configurations on \textbf{\benchmarkSmall}.
}
\label{table:lp-small}
\resizebox{0.99\textwidth}{!}{
\begin{tabular}{ l rrrrr } 
\toprule
 & RESCAL & TransE & ComplEx & ConvE & \update{TuckER} \\ 
\toprule
Best validation MRR & $0.4076$ & $0.3602$ & $0.4752$ & $0.4639$ & \update{$0.4574$} \\
Embedding size & $512$ & $512$ & $512$ & $256$ & \update{$512$}  \\ 
Training type & 1vsAll & NegSamp & 1vsAll & 1vsAll & \update{KvsAll} \\ 
\quad Reciprocal & No & Yes & Yes & Yes & \update{Yes} \\ 
\quad \# head samples (NegSamp) & - & $2$ & - & - & \update{-} \\ 
\quad \# tail samples (NegSamp) & - & $56$ & - & - & \update{-} \\ 
\quad Label smoothing (KvsAll) & - & - & - & -  & \update{$0.0950$} \\ 
Loss & CE & CE & CE & CE & \update{CE} \\ 
\quad Margin (MR) & - & - & - & - & \update{-} \\ 
\quad $\ell_p$ norm (TransE) & - & $2$ & - & - & \update{-} \\ 
Optimizer & Adagrad & Adagrad & Adam & Adagrad & \update{Adagrad} \\ 
\quad Batch size & $128$ & $128$ & $1024$ & $512$ & \update{$256$} \\ 
\quad Learning rate & $0.0452$ & $0.0412$ & $0.0003$ & $0.0117$ &  \update{$0.0145$} \\ 
\quad LR scheduler patience & $7$ & $6$ & $7$ & $3$ & \update{$1$} \\ 
$\ell_p$ regularization & $3$ & $2$ & None & $3$ & \update{$1$} \\ 
\quad Entity embedding weight & $2.18 \times 10^{-10}$ & $1.32 \times 10^{-7}$ & $9.58 \times 10^{-13}$ & $3.11 \times 10^{-15}$ & \update{$3.47 \times 10^{-15}$} \\ 
\quad Relation embedding weight & $3.37 \times 10^{-14}$ & $3.72 \times 10^{-18}$ & $0.0229$ & $4.68 \times 10^{-9}$ & \update{$3.43 \times 10^{-14}$} \\ 
\quad Frequency weighting & False & False & True & True & \update{True} \\ 
Embedding normalization (TransE) & & & & &  \\ 
\quad Entity & - & No & - & -  & \update{-} \\ 
\quad Relation & - & No & - & -  & \update{-} \\ 
Dropout & & & &  & \\ 
\quad Entity embedding & $0.0$ & $0.0$ & $0.0793$ & $0.0$  & \update{$0.1895$} \\ 
\quad Relation embedding & $0.0804$ & $0.0$ & $0.0564$ & $0.0$ & \update{$0.0$} \\ 
\quad Feature map (ConvE) & - & - & -& $0.2062$ & \update{-} \\ 
\quad Projection (ConvE) & - & - & - & $0.1709$ & \update{-} \\ 
Embedding initialization & Normal & XvNorm & XvNorm & XvNorm & \update{XvNorm} \\ 
\quad Stdev (Normal) & $0.0622$ & - & - & - & \update{-} \\ 
\quad Interval (Unif) & - & - & - & - & \update{-} \\ 
\quad Gain (XvNorm) & - & $1.0$ & $1.0$ & $1.0$ & \update{$1.0$} \\ 
\quad Gain (XvUnif) & - & - & - & - & \update{-} \\ 
\bottomrule
\end{tabular}

}
\end{table*}

\begin{table*}[t!]
\centering
\caption{Best \textbf{link prediction} hyperparameter configurations on \textbf{\benchmarkMed}.
}
\label{table:lp-medium}
\resizebox{0.99\textwidth}{!}{
\begin{tabular}{ l rrrrr } 
\toprule
 & RESCAL & TransE & ComplEx & ConvE & \update{TuckER} \\ 
\toprule
Best validation MRR & $0.3173$ & $0.2993$ & $0.3351$ & $0.3146$ & \update{$0.3253$}  \\
Embedding size & $256$ & $512$ & $512$ & $512$ & \update{$512$} \\ 
Training type & 1vsAll & NegSamp & KvsAll & NegSamp & \update{KvsAll} \\ 
\quad Reciprocal & Yes & Yes & Yes & Yes &  \update{Yes} \\ 
\quad \# head samples (NegSamp) & - & $2$ & - & $381$ & \update{-} \\ 
\quad \# tail samples (NegSamp) & - & $56$ & - & $751$ & \update{-} \\ 
\quad Label smoothing (KvsAll) & - & - & $0.2081$ & - & \update{$0.0950$} \\ 
Loss & CE & CE & CE & CE & \update{CE} \\ 
\quad Margin (MR) & - & - & - & - & \update{-} \\ 
\quad $\ell_p$ norm (TransE) & - & $2$ & - & - & \update{-} \\ 
Optimizer & Adagrad & Adagrad & Adagrad & Adagrad & \update{Adagrad} \\ 
\quad Batch size & $256$ & $128$ & $1024$ & $128$ & \update{$256$} \\ 
\quad Learning rate & $0.0695$ & $0.0412$ & $0.2557$ & $0.0024$ & \update{$0.0145$} \\ 
\quad LR scheduler patience & $8$ & $6$ & $6$ & $9$ & \update{$1$} \\ 
$\ell_p$ regularization & $2$ & $2$ & $3$ & $1$ & \update{$1$} \\ 
\quad Entity embedding weight & $9.56 \times 10^{-7}$ & $1.32 \times 10^{-7}$ & $1.34 \times 10^{-10}$ & $1.37 \times 10^{-10}$ & \update{$3.47 \times 10^{-15}$} \\ 
\quad Relation embedding weight & $2.56 \times 10^{-17}$ & $3.72 \times 10^{-18}$ & $6.38 \times 10^{-16}$ & $4.72 \times 10{-10}$ & \update{$3.4 \times 10^{-14}$} \\ 
\quad Frequency weighting & False & False & True & True & \update{True} \\ 
Embedding normalization (TransE) & & & & &  \\ 
\quad Entity & - & No & - & - & \update{-} \\ 
\quad Relation & - & No & - & - & \update{-} \\ 
Dropout & & & & &  \\ 
\quad Entity embedding & $0.0$ & $0.0$ & $0.1196$ & $0.0$ & \update{$0.1895$} \\ 
\quad Relation embedding & $0.0$ & $0.0$ & $0.3602$ & $0.0348$ & \update{$0.0$} \\ 
\quad Feature map (ConvE) & - & - & - & $0.3042$ & \update{-} \\ 
\quad Projection (ConvE) & - & - & - & $0.2343$ & \update{-} \\ 
Embedding initialization & XvUnif & XvUnif & Unif & XvNorm & \update{XvNorm} \\ 
\quad Stdev (Normal) & - & - & - & - & \update{-} \\ 
\quad Interval (Unif) & - & - & $-0.8133$ & - & \update{-} \\ 
\quad Gain (XvNorm) & - & - & - & $1.0$ & \update{$1.0$} \\ 
\quad Gain (XvUnif) & $1.0$ & $1.0$ & - & - & \update{-} \\ 
\bottomrule
\end{tabular}
}
\end{table*}

\begin{table*}[t!]
\centering
\caption{Best \textbf{link prediction} hyperparameter configurations on \textbf{\benchmarkLarge}.
}
\label{table:lp-large}
\resizebox{0.99\textwidth}{!}{
\begin{tabular}{ l rrrrr } 
\toprule
 & RESCAL & TransE & ComplEx & ConvE  & \update{TuckER} \\ 
\toprule
Best validation MRR & $0.3030$ & $0.1871$ & $0.2943$ & $0.3010$ & \update{$0.3091$} \\
Embedding size & $128$ & $128$ & $128$ & $256$ & \update{$256$} \\ 
Training type & 1vsAll & NegSamp & 1vsAll & 1vsAll  & \update{1vsAll} \\ 
\quad Reciprocal & No & Yes & Yes & Yes & \update{No} \\ 
\quad \# head samples (NegSamp) & - & $209$ & - & - & \update{-} \\ 
\quad \# tail samples (NegSamp) & - & $2$ & - & - & \update{-} \\ 
\quad Label smoothing (KvsAll) & - & - & - & -  & \update{-} \\ 
Loss & CE & CE & CE & CE  & \update{CE} \\ 
\quad Margin (MR) & - & - & - & -  & \update{-} \\ 
\quad $\ell_p$ norm (TransE) & - & $2$ & - & - & \update{-} \\ 
Optimizer & Adagrad & Adam & Adagrad & Adagrad & \update{Adagrad} \\ 
\quad Batch size & $1024$ & $128$ & $1024$ & $256$ & \update{$512$} \\ 
\quad Learning rate & $0.2651$ & $0.0009$ & $0.2651$ & $0.0329$ & \update{$0.0196$} \\ 
\quad LR scheduler patience & $7$ & $9$ & $7$ & $1$ & \update{$4$} \\ 
$\ell_p$ regularization & $2$ & $2$ & $2$ & $1$ & \update{$2$} \\ 
\quad Entity embedding weight & $2.01 \times 10^{-16}$ & $7.98 \times 10^{-14}$ & $2.01 \times 10^{-16}$ & $6.10 \times 10^{-16}$ & \update{$8.06 \times 10^{-11}$} \\ 
\quad Relation embedding weight & $3.52 \times 10^{-13}$ & $3.42 \times 10^{-9}$ & $3.52 \times 10^{-13}$ & $1.03 \times 10^{-16}$ & \update{$7.19 \times 10^{-19}$} \\ 
\quad Frequency weighting & True & False & True & True & \update{True} \\ 
Embedding normalization (TransE) & & & & &  \\ 
\quad Entity & - & No & - & - & \update{-} \\ 
\quad Relation & - & No & - & - & \update{-} \\ 
Dropout & & & & &  \\ 
\quad Entity embedding & $0.0$ & $0.0$ & $0.0$ & $0.0064$ & \update{$0.1606$} \\ 
\quad Relation embedding & $0.0$ & $0.0$ & $0.0$ & $0.0$ & \update{$0.0857$} \\ 
\quad Feature map (ConvE) & - & - & - & $0.1530$ & \update{-} \\ 
\quad Projection (ConvE) & - & - & - & $0.4192$ & \update{-} \\ 
Embedding initialization & Normal & Unif & Normal & XvNorm  & \update{Normal} \\ 
\quad Stdev (Normal) & $0.0169$ & - & $0.0169$ & -  & \update{$0.0002$} \\ 
\quad Interval (Unif) & - & $-0.4464$ & & -  & \update{-} \\ 
\quad Gain (XvNorm) & - & - & & $1.0$  & \update{-} \\ 
\quad Gain (XvUnif) & - & - & & -  & \update{-} \\ 
\bottomrule
\end{tabular}
}
\end{table*}

\begin{table*}[t!]
\centering
\caption{Best \textbf{triple classification} hyperparameter configurations on \textbf{\benchmarkSmall{}} (\textbf{hard negatives}).
}
\label{table:tc-small}
\resizebox{0.99\textwidth}{!}{
\begin{tabular}{ l rrrrr } 
\toprule
 & RESCAL & TransE & ComplEx & ConvE  & \update{TuckER} \\ 
\toprule
Best validation accuracy & \update{$0.8571$} & \update{$0.8511$} & \update{$0.8558$} & \update{$0.8607$} & \update{$0.8596$} \\ 
Embedding size & \update{See Tab.~\ref{table:lp-small}} & \update{See Tab.~\ref{table:lp-small}} & \update{See Tab.~\ref{table:lp-small}} &  \update{$512$} &  \update{See Tab.~\ref{table:lp-small}} \\ 
Training type & \update{1vsAll} & NegSamp & \update{1vsAll} & 1vsAll & \update{KvsAll} \\ 
\quad Reciprocal & \update{See Tab.~\ref{table:lp-small}} & \update{See Tab.~\ref{table:lp-small}} & \update{See Tab.~\ref{table:lp-small}} & \update{Yes}  & \update{See Tab.~\ref{table:lp-small}} \\ 
\quad \# head samples (NegSamp) & - & \update{See Tab.~\ref{table:lp-small}} & - & -  & \update{-} \\ 
\quad \# tail samples (NegSamp) & - & \update{See Tab.~\ref{table:lp-small}} & - & -  & \update{-} \\ 
\quad Label smoothing (KvsAll) & -  & - & - & -  & \update{-} \\ 
Loss & CE & \update{CE} & CE & \update{BCE}  & \update{CE} \\ 
\quad Margin (MR) & - & \update{-} & - & -  & \update{-} \\ 
\quad $\ell_p$ norm (TransE) & - & \update{See Tab.~\ref{table:lp-small}} & - & -  & \update{-} \\ 
Optimizer & \update{See Tab.~\ref{table:lp-small}} & \update{See Tab.~\ref{table:lp-small}} & \update{See Tab.~\ref{table:lp-small}} & \update{Adagrad}  & \update{See Tab.~\ref{table:lp-small}} \\ 
\quad Batch size & \update{See Tab.~\ref{table:lp-small}} & \update{See Tab.~\ref{table:lp-small}} & \update{See Tab.~\ref{table:lp-small}} & \update{$256$}  & \update{See Tab.~\ref{table:lp-small}}  \\ 
\quad Learning rate & \update{See Tab.~\ref{table:lp-small}} & \update{See Tab.~\ref{table:lp-small}} & \update{See Tab.~\ref{table:lp-small}} & \update{$0.0263$}  & \update{See Tab.~\ref{table:lp-small}} \\ 
\quad LR scheduler patience & \update{See Tab.~\ref{table:lp-small}} & \update{See Tab.~\ref{table:lp-small}} & \update{See Tab.~\ref{table:lp-small}} &  \update{$7$} & \update{See Tab.~\ref{table:lp-small}} \\ 
$\ell_p$ regularization & \update{See Tab.~\ref{table:lp-small}} & \update{See Tab.~\ref{table:lp-small}} & \update{See Tab.~\ref{table:lp-small}} & \update{$2$} & \update{See Tab.~\ref{table:lp-small}}  \\ 
\quad Entity embedding weight & \update{See Tab.~\ref{table:lp-small}} & \update{See Tab.~\ref{table:lp-small}} & \update{See Tab.~\ref{table:lp-small}} &  \update{$9.62 \times 10^{-6}$}  & \update{See Tab.~\ref{table:lp-small}}\\ 
\quad Relation embedding weight & \update{See Tab.~\ref{table:lp-small}} & \update{See Tab.~\ref{table:lp-small}} & \update{See Tab.~\ref{table:lp-small}} & \update{$1.34 \times 10^{-12}$}  & \update{See Tab.~\ref{table:lp-small}} \\ 
\quad Frequency weighting & \update{See Tab.~\ref{table:lp-small}} & \update{See Tab.~\ref{table:lp-small}} & \update{See Tab.~\ref{table:lp-small}} & \update{False}  &  \update{See Tab.~\ref{table:lp-small}} \\ 
Embedding normalization (TransE) & & & &  & \\ 
\quad Entity & - & \update{See Tab.~\ref{table:lp-small}} & - & -  & \update{-} \\ 
\quad Relation & - & \update{See Tab.~\ref{table:lp-small}} & - & -  & \update{-} \\ 
Dropout & & & &  & \\ 
\quad Entity embedding & \update{See Tab.~\ref{table:lp-small}} & \update{See Tab.~\ref{table:lp-small}}  & \update{See Tab.~\ref{table:lp-small}} & \update{$0.1620$}  & \update{See Tab.~\ref{table:lp-small}} \\ 
\quad Relation embedding & \update{See Tab.~\ref{table:lp-small}} & \update{See Tab.~\ref{table:lp-small}} & \update{See Tab.~\ref{table:lp-small}} & \update{$0.0031$}  & \update{See Tab.~\ref{table:lp-small}} \\ 
\quad Feature map (ConvE) & - & - & - & \update{$0.0682$}   & \update{-} \\ 
\quad Projection (ConvE) & - & - & - &  \update{$0.2375$} &  \update{-} \\ 
Embedding initialization & \update{See Tab.~\ref{table:lp-small}} & \update{See Tab.~\ref{table:lp-small}} & \update{See Tab.~\ref{table:lp-small}} & \update{Normal}  & \update{See Tab.~\ref{table:lp-small}} \\ 
\quad Stdev (Normal) & \update{See Tab.~\ref{table:lp-small}} & \update{See Tab.~\ref{table:lp-small}} & \update{See Tab.~\ref{table:lp-small}} & \update{$0.0006$} & \update{See Tab.~\ref{table:lp-small}}\\ 
\quad Interval (Unif) & \update{See Tab.~\ref{table:lp-small}} & \update{See Tab.~\ref{table:lp-small}} & \update{See Tab.~\ref{table:lp-small}} & -   & \update{See Tab.~\ref{table:lp-small}}\\ 
\quad Gain (XvNorm) & \update{See Tab.~\ref{table:lp-small}} & \update{See Tab.~\ref{table:lp-small}} & \update{See Tab.~\ref{table:lp-small}} &  -  & \update{See Tab.~\ref{table:lp-small}} \\ 
\quad Gain (XvUnif) & \update{See Tab.~\ref{table:lp-small}} & \update{See Tab.~\ref{table:lp-small}} & \update{See Tab.~\ref{table:lp-small}} &  -  & \update{See Tab.~\ref{table:lp-small}} \\ 
\bottomrule
\end{tabular}

}
\end{table*}

\begin{table*}[t!]
\centering
\caption{Best \textbf{triple classification} hyperparameter configurations on \textbf{\benchmarkMed{}} (\textbf{hard negatives}).
}
\label{table:tc-medium}
\resizebox{0.99\textwidth}{!}{
\begin{tabular}{ l rrrrr } 
\toprule
 & RESCAL & TransE & ComplEx & ConvE  & TuckER \\ 
\toprule
Best validation accuracy & \update{$0.8232$} & \update{$0.8002$} & \update{$0.8267$} & \update{$0.8292$}  & \update{$0.8267$} \\
Embedding size & $512$ & \update{See Tab.~\ref{table:lp-medium}} & $512$ & $512$  & \update{See Tab.~\ref{table:lp-medium}} \\ 
Training type & KvsAll & NegSamp & KvsAll & KvsAll  & KvsAll \\ 
\quad Reciprocal & Yes & \update{See Tab.~\ref{table:lp-medium}} & Yes & Yes  & \update{See Tab.~\ref{table:lp-medium}} \\ 
\quad \# head samples (NegSamp) & - & \update{See Tab.~\ref{table:lp-medium}} & - & -  & \update{-} \\ 
\quad \# tail samples (NegSamp) & - & \update{See Tab.~\ref{table:lp-medium}} & - & -  & \update{-} \\ 
\quad Label smoothing (KvsAll) & $0.0949$ & - & $0.2081$ & $0.0847$  & \update{-} \\ 
Loss & CE & \update{CE} & CE & CE  & \update{CE} \\ 
\quad Margin (MR) & - & \update{-} & - & -  & \update{-} \\ 
\quad $\ell_p$ norm (TransE) & - & \update{See Tab.~\ref{table:lp-medium}} & - & -  & \update{-} \\ 
Optimizer & Adagrad & \update{See Tab.~\ref{table:lp-medium}} & Adagrad & Adagrad  & \update{See Tab.~\ref{table:lp-medium}} \\ 
\quad Batch size & $256$ & \update{See Tab.~\ref{table:lp-medium}} & $1024$ & $1024$  & \update{See Tab.~\ref{table:lp-medium}} \\ 
\quad Learning rate & $0.0144$ & \update{See Tab.~\ref{table:lp-medium}} & $0.2557$ & $0.0378$  & \update{See Tab.~\ref{table:lp-medium}} \\ 
\quad LR scheduler patience & $1$ & \update{See Tab.~\ref{table:lp-medium}} & $6$ & $6$  & \update{See Tab.~\ref{table:lp-medium}} \\ 
$\ell_p$ regularization & $1$ & \update{See Tab.~\ref{table:lp-medium}} & $3$ & $3$ & \update{See Tab.~\ref{table:lp-medium}} \\ 
\quad Entity embedding weight & $3.47 \times 10^{-15}$ & \update{See Tab.~\ref{table:lp-medium}} & $1.34 \times 10^{-10}$ & $1.03\times 10^{-16}$  & \update{See Tab.~\ref{table:lp-medium}} \\ 
\quad Relation embedding weight & $3.43 \times 10^{-14}$ & \update{See Tab.~\ref{table:lp-medium}} & $6.38 \times 10^{-16}$  & $0.0052$  & \update{See Tab.~\ref{table:lp-medium}} \\ 
\quad Frequency weighting & True & \update{See Tab.~\ref{table:lp-medium}} & True & True  & \update{See Tab.~\ref{table:lp-medium}} \\ 
Embedding normalization (TransE) & & & &  & \\ 
\quad Entity & - & \update{See Tab.~\ref{table:lp-medium}} & - & -  & \update{-} \\ 
\quad Relation & - & \update{See Tab.~\ref{table:lp-medium}} & - & -  & \update{-} \\ 
Dropout & & & &  & \\ 
\quad Entity embedding & $0.1895$ & \update{See Tab.~\ref{table:lp-medium}} & $0.1196$ & $0.4828$  & \update{See Tab.~\ref{table:lp-medium}} \\ 
\quad Relation embedding & $0.0$ & \update{See Tab.~\ref{table:lp-medium}} & $0.3602$ & $0.0$  & \update{See Tab.~\ref{table:lp-medium}} \\ 
\quad Feature map (ConvE) & - & - & - & $0.2649$  & \update{-} \\ 
\quad Projection (ConvE) & - & - & - & $0.2790$  & \update{-} \\ 
Embedding initialization & XvNorm & \update{See Tab.~\ref{table:lp-medium}} & Unif & XvUnif  & \update{See Tab.~\ref{table:lp-medium}} \\ 
\quad Stdev (Normal) & - & \update{See Tab.~\ref{table:lp-medium}} & - & -  & \update{See Tab.~\ref{table:lp-medium}} \\ 
\quad Interval (Unif) & - & \update{See Tab.~\ref{table:lp-medium}}  & $-0.8133$ & -  & \update{See Tab.~\ref{table:lp-medium}} \\ 
\quad Gain (XvNorm) & $1.0$ & \update{See Tab.~\ref{table:lp-medium}} & - & -  & \update{See Tab.~\ref{table:lp-medium}} \\ 
\quad Gain (XvUnif) & - & \update{See Tab.~\ref{table:lp-medium}} & - & $1.0$  & \update{See Tab.~\ref{table:lp-medium}} \\ 
\bottomrule
\end{tabular}

}
\end{table*}

\end{document}